%% file: coling.tex
\pdfoutput=1

\documentclass[11pt]{article}

\usepackage[final]{coling}

\usepackage{times}
\usepackage{latexsym}

\usepackage[T1]{fontenc}

\usepackage[utf8]{inputenc}

\usepackage{microtype}

\usepackage{inconsolata}

\usepackage{graphicx}

\usepackage{graphicx}
\usepackage{multirow}
\usepackage{booktabs}
\usepackage{pifont}
\newcommand{\cmark}{\ding{51}}
\usepackage{rotating}
\usepackage{xcolor}
\usepackage[normalem]{ulem}
\usepackage{float}
\usepackage{balance}

\title{STAND-Guard: A Small Task-Adaptive Content Moderation Model
\\
\small \centering\textcolor{red}{Disclaimer: The paper may contain usage of potentially offensive, sensitive or mature content.}}

\author{
 \textbf{Minjia Wang\thanks{\, Work done during the internship at Microsoft.}\textsuperscript{1, 2}},
 \textbf{Pingping Lin\textsuperscript{1}},
 \textbf{Siqi Cai\textsuperscript{1, 3}},
 \textbf{Shengnan An\textsuperscript{1, 4}},
\\
 \textbf{Shengjie Ma\textsuperscript{1, 5}},
 \textbf{Zeqi Lin\textsuperscript{1}},
 \textbf{Congrui Huang\textsuperscript{1}},
 \textbf{Bixiong Xu \textsuperscript{1}}
\\
\\
 \textsuperscript{1}Microsoft, 
 \textsuperscript{2}Harvard University, 
 \textsuperscript{3}Peking University, \\
 \textsuperscript{4}Xi'an Jiaotong University, 
  \textsuperscript{5}University of Illinois Urbana-Champaign
\\
 \textsuperscript{1} \small \texttt{
   \{v-minjiawang, pinlin, v-siqicai, t-shengnanan,v-shengjiema, Zeqi.Lin, }\\
   \small\texttt{ conhua, bix\}@microsoft.com} 
   \textsuperscript{2}\small \texttt{minjiawang@g.harvard.edu},\\
   \textsuperscript{3}\small \texttt{2201210579@pku.edu.cn}, 
   \textsuperscript{4}
   \small \texttt{an1006634493@stu.xjtu.edu.cn}, 
    \textsuperscript{5}\small \texttt{sm138@illinois.edu}
 }

\begin{document}
\maketitle
\input{src/sec0-abstract}
\input{src/sec1-introduction}
\input{src/sec2-related-work}
\input{src/sec3-methodology}
\input{src/sec4-experimental-setup}
\input{src/sec5-results}
\input{src/sec6-conclusion}

\bibliography{coling}

\appendix
\input{src/sec8-appendix}

\end{document}

%% file: src/sec0-abstract.tex
\begin{abstract}
Content moderation, the process of reviewing and monitoring the safety of generated content, is important for development of welcoming online platforms and responsible large language models.
Content moderation contains various tasks, each with its unique requirements tailored to specific scenarios.
Therefore, it is crucial to develop a model that can be easily adapted to novel or customized content moderation tasks accurately without extensive model tuning.
This paper presents \textsc{STAND-Guard}, a Small Task-Adaptive coNtent moDeration model.
The basic motivation is: by performing instruct tuning on various content moderation tasks, we can unleash the power of small language models (SLMs) on unseen (out-of-distribution) content moderation tasks.
We also carefully study the effects of training tasks and model size on the efficacy of cross-task fine-tuning mechanism.
Experiments demonstrate STAND-Guard is comparable to GPT-3.5-Turbo across over 40 public datasets, as well as proprietary datasets derived from real-world business scenarios. Remarkably, STAND-Guard achieved nearly equivalent results to GPT-4-Turbo on unseen English binary classification tasks.
\end{abstract}
 

%% file: src/sec1-introduction.tex
\section{Introduction}

Ensuring content safety is essential for online communities and social media platforms to maintain a friendly communication environment \cite{arora2023detecting}. With the rapid development of large language models (LLMs), content moderation has also become crucial for service providers to preserve model quality and safeguard user interactions \cite{markov2023holistic}.

Industries are developing automated content moderation algorithms to ensure online content safety and integrity. Recent advancements in deep learning have established supervised training of lightweight classifiers as a typical paradigm \cite{markov2023holistic}. This approach provides a low-cost and efficient way to filter undesired content. However, it also faces challenges, such as aligning sufficient training data with evolving community policies, and updating human reviewers on new harmful categories. Even with adequate training data, these classifiers, which are trained on fixed and labeled datasets for a specific task, may still struggle to cope with the diversity and complexity of textual content. They are inflexible to transfer to out-of-distribution tasks. On the other hand, while the success of generative LLMs like GPT-4 motivates their use in content moderation\footnote{https://openai.com/index/using-gpt-4-for-content-moderation/}, this approach has limitations. The substantial cost of hosting LLMs and the risk of a single LLM's vulnerabilities being exploited by malicious actors, present significant challenges. All these methods lack practicality in real-world business scenarios.

Thus, we need to build a content moderation model which is much smaller and cheaper than those LLMs, like GPT-4, but still have enough domain knowledge and adaptability to handle new tasks with or without few-shot examples. Now this approach raises several questions:
1) How well can the moderation model cope with out-of-distribution data that may occur in real-world scenarios?
2) How can we obtain data to generate the model for content moderation, given that human annotations are costly and scarce?
3) How much data and how many tasks do we need to train the model effectively? Is there a trade-off between the number of tasks and the model's performance, or does more data always lead to better results? 

Nowadays, small language models (SLMs), like Mistral-7B \cite{jiang2023mistral}, Gemma \cite{team2024gemma} and Phi-3 \cite{abdin2024phi} have shown impressive performance or even superiority in some domains. Additionally, hosting a SLM requires fewer resources. These inspire us to address the aforementioned challenges by fine-tuning these SLMs~\cite{ma2023adapting, zhang2024efficient, umanskyenhancing}. We propose a cross-task fine-tuning method specifically tailored for SLMs, focusing on content moderation domain. Our work contributes in three significant ways:
\vspace{-0.1cm}
\begin{itemize}
\vspace{-0.1cm}
    \item We present a methodology, namely cross-task fine-tuning, to fine-tune SLMs for novel content moderation tasks, specifically for out-of-distribution data. 
    \vspace{-0.25cm}
    \item We categorize public datasets into different tasks and use them for cross-task fine-tuning in content moderation. Through various experiments, we demonstrate the potential of cross-task fine-tuning a business model with public datasets, making it highly practical. 
    \vspace{-0.25cm}
    \item We develop a unified task-adaptive model, STAND-Guard, through cross-task fine-tuning. We evaluate the model on both public and proprietary business datasets. The model surpasses GPT-3.5-Turbo on in-distribution data, and performs on par with GPT-3.5-Turbo on out-of-distribution tasks. Notably, STAND-Guard achieves comparable results to GPT-4-Turbo on unseen (out-of-distribution) English binary classification tasks.
\end{itemize}

%% file: src/sec2-related-work.tex
\section{Related Work}
\vspace{-0.15cm}
Advancements in Large and Small Language Models (LLMs and SLMs) have made them viable for various tasks, including content moderation. These models can be utilized through two main methods: prompting and fine-tuning.

\textbf{Prompting} involves providing the LLM/SLM with a specific query or instruction, which it then uses to generate a response.
In terms of content moderation, the prompt generally incorporates the moderation guidelines, along with the content subject to review~\cite{kolla2024llm}. 
There are many prompting strategies~\cite{guo2023investigation, franco2023analyzing, kumar2024watch,zhang2023interpretable,zhang2024efficient}, nevertheless, crafting prompts that accurately reflect the moderation guidelines while also enhancing the performance of language models demands significant human intervention and computational resources. Therefore, we do not primarily focus on prompt engineering in this study. To ensure a fair comparison, we will utilize the same prompt across various baselines and models.

\textbf{Fine-tuning}, on the other hand, involves adjusting the LLM's or SLM's parameters to better suit a particular task such as content moderation.
Some methods update all model parameters~\cite{ghosh2024aegis}, which is a resource-intensive process due to the substantial size of language models. Consequently, more efficient alternative methods have been developed, which modify only a subset of parameters. Techniques under this category include adding task-specific layers~\cite{wullach2021fight, markov2023holistic,houlsby2019parameter,sen2024hatetinyllm}, LoRA~\citep{hu2021lora,ma2023adapting} and prompt-tuning~\cite{li2021prefix, he2023you,liu2022p, markov2023holistic,qiao2024scaling, lester2021power,yuan2024rigorllm}. However, many of these approaches are restricted to specific content moderation tasks or undesired categories~\citep{markov2023holistic, guo2023investigation, kolla2024llm}, limiting their capability to generalize to new tasks and categories. 

%% file: src/sec3-methodology.tex
\section{Methodology}
\vspace{-0.15cm}
\subsection{Cross-task fine-tuning}
\vspace{-0.15cm}
A \textit{task} is an annotation process that determines if content requires modification, or identifies the types of harm or targeted groups involved. Each task is inherently linked with a guideline that outlines the procedure for the annotation process.

It is well-established that fine-tuning boosts the performance of \textit{in-distribution tasks} ~\cite{ma2023adapting}, i.e., tasks encountered during fine-tuning. However, the computational intensity required to fine-tune a model for every  task is considerable. Moreover, it is not feasible in actual business scenarios. The question then arises: can fine-tuned models sustain or even enhance their performance when dealing with \textit{out-of-distribution tasks}, i.e., tasks not present during fine-tuning?

To answer this question, we propose cross-task fine-tuning, which is designed to adapt the content moderation model to new tasks quickly during inference, and to improve out-of-distribution task performance by incorporating additional tasks into the training set during fine-tuning.

\vspace{-0.3cm}
\subsection{Building the training set}
\label{sec:methodology-building-training-set}
When constructing the training set, some practical questions arise. For instance, the selection of tasks and the quantity of tasks to be included in the training set. Our primary goal is to design a training set that is as minimal as possible, yet still yields a substantial increase in performance.

To this end, we first group content moderation tasks into categories and subcategories. Then, we curate a compact training set that encompasses all the subcategories, utilizing only a single private dataset and two public datasets. 

Based on ~\citet{wang2023not}, we developed a two-level taxonomy for content moderation tasks, categorizing them into 4 primary categories and 8 subcategories. The 4 primary categories are \textit{Malicious Actions}, \textit{Discrimination / Exclusion / Toxicity / Hateful / Offensive}, \textit{Information Hazards} and \textit{Misinformation Harms}. Please refer to Appendix \ref{sec:categories} for detailed definitions. 

\vspace{-0.2cm}
\subsection{Fine-tuning models}
\label{fine-tuning-models}

\paragraph{Problem definition} Let's consider a content moderation task characterized by a guideline $G$, and a corresponding dataset represented as $\mathcal{D} = \big( G, \{x_i, y_i\}_{i=0}^{N-1} \big)$. $G$ is the moderation guideline in a human-readable format that describes the annotation standards, and specifies the output format of language models. $x_i$ signifies the input content for the sample indexed at $i$, and $y_i$ which falls in the set $\{0, 1, ..., K_{\mathcal{D}} - 1\}$ indicates the respective ground truth label. Given $\{G, x_i\}$, the goal for language models is to predict $y_i$.

\vspace{-0.15cm}
\paragraph{Guideline generation}
A guideline $G$ comprised of two parts: 1) Definitions of the undesired content. For public datasets, these are extracted from the dataset description or the original paper if available; otherwise, they are generated by GPT-4-Turbo (see Appendix~\ref{sec:definition-generation} for details). For private datasets, our internal guidelines are used. 2) The evaluation process, which specifies the label set (e.g., binary classification or multi-class classification) , factors to consider during evaluation, and the expected output format.

\vspace{-0.15cm}
\paragraph{Fine-tuning with QLoRA}

We chose to fine-tune SLMs based on QLoRA~\cite{dettmers2024qlora}, which combines Quantization and Low-Rank Adapters to allow for efficient fine-tuning.
The input during fine-tuning is the guideline $G$ and the content $x_i$ to be reviewed. An illustrative example of these input prompts can be found in Appendix~\ref{sec:example-prompt}. Note that we do not primarily focus on prompt engineering in this work, and just utilize the same prompt format across various tasks, baselines and models. The expected output is "Label: $y_i$" for each training sample.

%% file: src/sec4-experimental-setup.tex
\section{Experimental Setup}
\vspace{-0.15cm}
\subsection{Implementation details}
STAND-Guard uses Mistral-7B \cite{jiang2023mistral} v0.1, a 7-billion-parameter model from Mistral AI, as backbone model. 
To assess the impact of the backbone model's size on the efficacy of cross-task fine-tuning, we compare the performance of STAND-Guard with models underpinned by different backbones: Phi-3-mini-128k-instruct \cite{abdin2024phi} (3.8 billion parameters) and Mixtral-8×7B \cite{jiang2024mixtral} v0.1.
They were fine-tuned, inferred and evaluated using the process and configuration detailed in Appendix \ref{sec:implementation-details}.

\vspace{-0.15cm}
\subsection{Baseline models}
\label{sec:experiment-models}
\vspace{-0.15cm}
We benchmark STAND-Guard against two sets of baseline models, namely task-specific models and general models. \textit{Task-specific models} are trained for specific tasks but do not accommodate the input of custom policies, including Perspective API and OpenAI Content Moderation API. \textit{General models} are designed to accept guideline as in-context input to steer the classification of the input text, including LlamaGuard, GPT-3.5-Turbo and GPT-4-Turbo. We provide a brief overview of these baselines in Appendix \ref{sec:baseline-models}.

\subsection{Data preparation}
\label{sec:dataset-stats}
We have collected data from related research as well as various public repositories, with a summary provided in Appendix \ref{sec:dataset-correlation}.
We have maintained the separation between the training and test sets for each dataset to ensure no overlap between them. The statistics of the training and evaluation datasets are presented in Tables \ref{tab:train-dataset-stats} and \ref{tab:test-dataset-stats}.

\vspace{-0.15cm}
\paragraph{Training dataset} As mentioned in Section \ref{sec:methodology-building-training-set},
to build a training set as minimal as possible,
only the following three datasets are used as the training material.

\emph{PKU-Alignment BeaverTails} and \emph{PKU-Alignment SafeRLHF} \cite{ji2024beavertails} are datasets for safety alignment in LLMs including helpfulness and harmlessness. The datasets comprises dozens of tasks. We utilize its data solely for safety assessment purposes and convert each task into a binary classification task.

\emph{Private} dataset is a collection curated from our business context, comprising texts that have been manually annotated for five distinct categories, including labels for sexual content, self-harm, violence, hate speech and jailbreak.

\vspace{-0.15cm}
\paragraph{Evaluation dataset}
We evaluate our methods against the two groups of models described in Section \ref{sec:experiment-models}.
For task-specific models, we adopt the approach of \citet{markov2023holistic} to conduct comparisons across only four datasets primarily associated with hate speech, offensive language, and toxicity. These datasets include the \emph{OpenAI Content Moderation dataset} (\emph{OpenAI CM} for short) \cite{markov2023holistic}, \emph{Jigsaw}\footnote{https://www.kaggle.com/c/jigsaw-toxic-comment-classification-challenge}, \emph{TweetEval} \cite{barbieri2020tweeteval}, and \emph{White Supremacist} \cite{de2018hate} as shown in Appendices \ref{sec:dataset-correlation} and \ref{sec:stats-of-test-tasks}. Due to the rate limits imposed by the Perspective API and OpenAI Content Moderation API, we sampled 5,000 entries from the entire Jigsaw dataset for our analysis. For general models that accommodate input based on custom guidelines, we conduct a comprehensive comparison across 42 datasets and 80 tasks.

%% file: src/sec5-results.tex
\vspace{-0.3cm}
\section{Results and Analysis}
\vspace{-0.15cm}
\subsection{In-distribution tasks}

\begin{table*}\scriptsize
  \centering
\begin{tabular}{llrrrr|r}
\toprule
Dataset &Task&LlamaGuard &GPT-3.5-Turbo &GPT-4-Turbo &Mistral-7B &\textbf{STAND-Guard} \\
 \midrule
PKU-Alignment BeaverTails &Animal Abuse &0.580 &0.341 &0.694 &0.438 &0.742 \\
 &Child Abuse &0.553 &0.176 &0.372 &0.325 &0.815 \\
 &Controversial Topics, Politics &0.034 &0.056 &0.043 &0.114 &0.446 \\
 &Discrimination, Stereotype &0.618 &0.348 &0.330 &0.456 &0.731 \\
 &Drug Abuse, Weapons &0.611 &0.457 &0.317 &0.419 &0.746 \\
 &Financial \& Property Crime &0.592 &0.521 &0.515 &0.539 &0.744 \\
 &Hateful \& Offensive Language &0.529 &0.328 &0.326 &0.254 &0.670 \\
 &Misinformation &0.037 &0.060 &0.066 &0.052 &0.082 \\
 &Non-Violent Unethical Behavior &0.207 &0.384 &0.418 &0.199 &0.655 \\
 &Privacy Violation &0.303 &0.177 &0.449 &0.288 &0.800 \\
 &Self Harm &0.522 &0.063 &0.078 &0.110 &0.727 \\
 &Sexually Explicit &0.537 &0.358 &0.580 &0.410 &0.667 \\
 &Terrorism, Organized Crime &0.067 &0.143 &0.097 &0.191 &0.196 \\
 &Violence &0.253 &0.672 &0.681 &0.397 &0.800 \\
PKU-Alignment   Safe-RLHF &Unsafe &0.580 &0.763 &0.818 &0.492 &0.871 \\
Private &Hate &0.700 &0.745 &0.697 &0.642 &0.827 \\
 &Self Harm &0.654 &0.573 &0.707 &0.556 &0.856 \\
 &Sexual &0.335 &0.660 &0.800 &0.010 &0.802 \\
 &Violence &0.324 &0.538 &0.719 &0.486 &0.745 \\
 \hline
\multicolumn{2}{c}{AVG} &0.423 &0.388 &0.458 &0.336 & 0.680 \\
\bottomrule
\end{tabular}
\vspace{-0.2cm}
  \caption{\label{tab:in-distribution-results}
    F1 scores on in-distribution tasks under zero-shot setting. Jailbreak in our private dataset does not have a test set.
  }
\end{table*}

Table~\ref{tab:in-distribution-results} presents the F1 scores on in-distribution tasks, from which we can draw three conclusions.

1) STAND-Guard, fine-tuned with cross-task learning, markedly surpasses the performance of the vanilla Mistral-7B, showcasing the effectiveness of fine-tuning on in-distribution tasks.

2) STAND-Guard outperforms GPT-4-Turbo, one of the most advanced LLMs, in content moderation. This indicates that specialized, fine-tuned smaller models can excel in specific tasks compared to a generic LLM. This insight implies that if training data is attainable in a business scenario, we can employ it for fine-tuning in order to achieve results comparable to those of GPT-4-Turbo. Appendix~\ref{sec:gpt4-error} includes an error analysis for GPT-4-Turbo. 

3) The F1 score for the same model varies widely across datasets, despite the use of a uniform guideline generation method. This variation reflects the intrinsic differences between tasks, as detailed in a case study in Appendix \ref{sec:exp-casestudy}.

\vspace{-0.2cm}
\subsection{Out-of-distribution tasks}

\begin{table*}[h!]\scriptsize
  \centering
\begin{tabular}{llrrrr|r}
\toprule
{Dataset} &{Task} &{LlamaGuard} &{GPT-3.5-Turbo} &{GPT-4-Turbo} &Mistral-7B &{\textbf{STAND-Guard}} \\
 \midrule
CallMeSexist &Sexism &0.145 &0.631 &0.639 &0.228 &0.724 \\
Civil-Comments &Insult &0.533 &0.674 &0.801 &0.599 &0.787 \\
 &Obscenity &0.028 &0.347 &0.346 &0.609 &0.179 \\
 &Severe Toxicity &0.481 &0.416 &0.141 &0.582 &0.530 \\
 &Sexually Explicit &0.016 &0.142 &0.029 &0.483 &0.134 \\
 &Threat &0.080 &0.376 &0.188 &0.454 &0.163 \\
 &Toxicity &0.494 &0.730 &0.474 &0.573 &0.759 \\
Commonsense Morality &Ethics &0.014 &0.809 &0.874 &0.711 &0.735 \\
CrowS-Pairs &Bias &0.675 &0.707 &0.778 &1.000 &1.000 \\
DecodingTrust &Stereotype &0.985 &0.943 &1.000 &1.000 &1.000 \\
DynaHate &Hate &0.150 &0.827 &0.834 &0.612 &0.673 \\
Exaggerated Safety &Safety &0.020 &0.882 &0.950 &0.038 &0.829 \\
HASOC   (English) &Hate, Offensive &0.038 &0.367 &0.576 &0.323 &0.679 \\
HateCheck &Hate &0.829 &0.949 &0.961 &0.814 &0.867 \\
HateEval &Hate &0.666 &0.005 &0.655 &0.639 &0.462 \\
HatemojiCheck &Hate &0.256 &0.825 &0.920 &0.774 &0.836 \\
HateXplain &Hate &0.788 &0.796 &0.820 &0.220 &0.782 \\
Jigsaw &Identity Hate &0.232 &0.111 &0.254 &0.021 &0.281 \\
 &Insult &0.501 &0.261 &0.447 &0.080 &0.416 \\
& Obscene &0.167 &0.322 &0.534 &0.084 &0.512 \\
 &Severe Toxic &0.112 &0.065 &0.141 &0.011 &0.085 \\
 &Threat &0.221 &0.030 &0.253 &0.006 &0.300 \\
 &Toxic &0.549 &0.359 &0.474 &0.127 &0.551 \\
OpenAI CM &Harassment &0.161 &0.255 &0.268 &0.077 &0.726 \\
 &Self Harm &0.000 &0.292 &0.630 &0.099 &0.928 \\
Reddit Content Moderation &Rule Moderation &0.002 &0.467 &0.445 &0.159 &0.126 \\
Scruples Anecdotes &Ethics &0.000 &0.445 &0.555 &0.061 &0.427 \\
Social   Bias Inference Corpus (SBIC) &Intentionally Offensive &0.707 &0.726 &0.722 &0.665 &0.709 \\
 &Potentially Offensive &0.738 &0.738 &0.731 &0.633 &0.728 \\
 &Sexually Offensive &0.509 &0.666 &0.641 &0.689 &0.195 \\
SWAD &Swear &0.007 &0.447 &0.551 &0.415 &0.539 \\
ToxiGen &Toxic &0.729 &0.779 &0.815 &0.497 &0.601 \\
TrustworthyLLM &Safety &0.225 &0.571 &0.874 &0.402 &0.590 \\
TweetEval &Hate &0.650 &0.686 &0.486 &0.308 &0.556 \\
 &Irony &0.061 &0.685 &0.780 &0.005 &0.685 \\
 &Offensive &0.381 &0.582 &0.525 &0.573 &0.682 \\
USElectionHate &Hate &0.392 &0.346 &0.504 &0.034 &0.392 \\
White Supremacist &Hate &0.503 &0.796 &0.711 &0.582 &0.739 \\
\hline
\multicolumn{2}{c}{AVG} &0.343 &0.528 &0.588 &0.400 &0.577
\\\bottomrule
\end{tabular}
\vspace{-0.2cm}
  \caption{\label{tab:out-of-distribution-results-en-binary}
     F1 scores on out-of-distribution tasks (binary classification, English data) under zero-shot setting.
  }
\end{table*}

\vspace{-0.1cm}
\subsubsection{Main results}
\vspace{-0.1cm}
Table \ref{tab:out-of-distribution-results-en-binary} presents the F1 scores for out-of-distribution English binary classification tasks across a broad spectrum of tasks. 

\vspace{-0.2cm}
\paragraph{STAND-Guard vs. vanilla models.} Upon examing Table \ref{tab:out-of-distribution-results-en-binary}, a conclusion emerges: the results highlight the performance improvements of the model that underwent fine-tuning on selected tasks when assessed against a wide array of out-of-distribution tasks, in comparison to the vanilla model. It also shows that the fine-tuned model has achieved considerable gains in relatively novel tasks such as \emph{irony detection}, \emph{harassment detection}, and \emph{toxicity detection}, surpassing the vanilla model's performance in these areas. These findings robustly endorse the efficacy of cross-task knowledge transfer, as it demonstrates the model's enhanced adaptability and generalization capabilities across various datasets and tasks.
\vspace{-0.2cm}
\paragraph{STAND-Guard vs. GPT models.} Additionally, the results from Table \ref{tab:out-of-distribution-results-en-binary} indicate that the model, fine-tuned via cross-task methods, not only exceeds the performance of the larger GPT-3.5-Turbo (0.528 $\rightarrow$ 0.577) but also exhibit small performance drop-off when compared to GPT-4-Turbo on binary classification tasks represented in English. 
\vspace{-0.2cm}
\paragraph{STAND-Guard vs. task-specific API models.} Table \ref{tab:out-of-distribution-results-hate} in Appendix \ref{sec:comparison-task-specific} presents the results obtained using Perspective API and OpenAI Content Moderation API on datasets concerning hate and offensive language, following \citet{markov2023holistic}'s methodology. It demonstrates that STAND-Guard is the only task-adaptive model that outstrips task-specific API models. 
Moreover, STAND-Guard not only achieves the best results among all baselines—including API models trained on datasets related to hate speech and offensive language—but it also outshines the performance of GPT-3.5-Turbo and GPT-4-Turbo.

It is important to acknowledge that tasks derived from different datasets might show inconsistencies, which can be attributed to nuanced differences in their underlying concepts or definitions.
We further explore the correlation between task semantic similarity and classification quality improvements on these tasks in Appendix \ref{sec:vis-task-similarity}.
\vspace{-0.1cm}
\subsubsection{Additional analysis}

We also conduct an in-depth examination of performance across \textbf{multi-lingual} tasks and \textbf{multi-class classification} tasks. The results are displayed in Tables \ref{tab:out-of-distribution-results-non-en} and \ref{tab:out-of-distribution-results-multiclass}, respectively.
Under such experimental settings, the overall performance of the fine-tuned model exhibits a reduction when compared to both GPT-3.5-Turbo and GPT-4-Turbo. This underlines the significance of maintaining a close alignment between the distribution of the training and testing data. This drop is due to fine-tuning with exclusively English binary classification data, highlighting the necessity for a more diverse training corpus to achieve optimal performance across varied linguistic contexts and task complexities. 
The fine-tuned model retains its \textbf{in-context learning} abilities, as illustrated in Appendix \ref{sec:in-context-learning}. When combined with Retrieval Augmented Generation (RAG) \cite{an2023context, hu2022context}, it is capable of attaining competitive performance on par with the zero-shot capabilities of GPT-4-Turbo.

\begin{table*}[h!]\scriptsize
\centering
\begin{tabular}{llrrrr}
\toprule
\multirow{2}{*}{Dataset}    &\multirow{2}{*}{Task}   &
\multirow{2}{*}{Mistral-7B} &STAND-Guard  &STAND-Guard       &\multirow{2}{*}{STAND-Guard}   \\
       &   & &(w/o Private) &(w/o Hate Offensive) &\\
\midrule
PKU-Alignment BeaverTails       &Animal Abuse   &0.438 &0.697 &0.716 &0.742 \\
       &Child Abuse    &0.325 &0.792 &0.857 &0.815 \\
       &Controversial Topics, Politics &0.114 &0.372 &0.450 &0.446 \\
       &Discrimination, Stereotype     &0.456 &0.734 &0.734 &0.731 \\
       &Drug Abuse, Weapons    &0.419 &0.745 &0.698 &0.746 \\
       &Financial \& Property Crime    &0.539 &0.742 &0.751 &0.744 \\
       &Hateful \& Offensive Language  &0.254 &0.663 &\textbf{0.571} &0.670 \\
       &Misinformation &0.052 &0.000 &0.051 &0.082 \\
       &Non-Violent Unethical Behavior &0.199 &0.633 &0.657 &0.655 \\
       &Privacy Violation      &0.288 &0.782 &0.791 &0.800 \\
       &Self Harm      &0.110 &0.710 &0.667 &0.727 \\
       &Sexually Explicit      &0.410 &0.634 &0.611 &0.667 \\
       &Terrorism, Organized Crime     &0.191 &0.089 &0.163 &0.196 \\
       &Violence       &0.397 &0.791 &0.796 &0.800 \\
PKU-Alignment   Safe-RLHF       &Unsafe&0.492 &0.846 &0.844 &0.871 \\
Private&Hate  &0.642 &\textbf{0.700} &\textbf{0.729} &0.827 \\
       &Self Harm      &0.556 &\textbf{0.813} &0.839 &0.856 \\
       &Sexual&0.010 &\textbf{0.639} &0.671 &0.802 \\
       &Violence       &0.486 &\textbf{0.743} &0.705 &0.745 \\
\hline
\multicolumn{2}{c}{AVG}     &0.336 &0.638 &0.647 &0.680  
\\\bottomrule
\end{tabular}
\vspace{-0.2cm}
  \caption{\label{tab:exp-ablation}
    Ablation study on training data. STAND-Guard (w/o Private): our proprietary datasets are excluded from the training set. STAND-Guard (w/o Hate Offensive): data related to hate speech and offensive content are removed. F1 scores on out-of-distribution tasks (i.e., tasks not in the training set) are in bold.}
\end{table*}

\vspace{-0.3cm}
\subsection{Ablation study}
\vspace{-0.25cm}
Table \ref{tab:exp-ablation} presents the results of our training data ablation study. We systematically removed portions of the training data, initially detailed in Section \ref{sec:dataset-stats}, to evaluate cross-task knowledge transfer. Two additional experiments were conducted: 
1) \textbf{STAND-Guard (w/o Private)}: excluded our proprietary datasets from the full training set.
2) \textbf{STAND-Guard (w/o Hate Offensive)}: removed data related to hate speech and offensive content from the full training set.
From Table \ref{tab:exp-ablation}, we can draw three conclusions:

1) Even without proprietary datasets, STAND-Guard (w/o Private) showed significant improvements over vanilla Mistral-7B on private datasets. This finding is encouraging as it suggests that individuals can align a language model with business-specific guidelines by fine-tuning it solely on publicly available content moderation datasets, rather than relying on business data. This approach has the potential to save substantial time and resources that would otherwise be spent on collecting human-labeled data for the training set.

2) Despite the absence of hate speech and offensive content data, STAND-Guard (w/o Hate Offensive) still surpasses vanilla Mistral-7B. This indicates that tasks not directly related to hate or offensive language can still contribute positively to the detection of such content. This outcome further validates the efficacy of cross-task fine-tuning.

3) The model that was fine-tuned using the complete training dataset either outperformed or matched the performance of the two models trained on partial datasets across all tasks. This suggests that the private dataset (or the hate detection dataset) is capable of transferring knowledge to virtually all tasks, or at the very least, does not diminish the quality of detection.

\vspace{-0.2cm}
\subsection{Influence of model size}

\begin{table}\scriptsize
\begin{tabular}{lrrr}
\toprule
\multirow{2}{*}{Dataset/Task} &Phi-3-mini &\multirow{2}{*}{STAND-Guard} &Mixtral-8x7B \\
     &(CT-FT)    &     &(CT-FT)     \\
\midrule
in-distribution       &0.581      &0.680&0.671\\
out-of-distribution   &0.488      &0.533&0.577 
\\\bottomrule
\end{tabular}
\vspace{-0.15cm}
\caption{\label{tab:f1-model-size}Average F1 scores for cross-task fine-tuned models of various sizes.}
\end{table}
We analyze the impact of model size on cross-task fine-tuning by comparing the performance of three backbone models of various sizes: Phi-3-mini (3.8B parameters), STAND-Guard (7B parameters), and Mixtral-8×7B (56B parameters). Table \ref{tab:f1-model-size} presents the average F1 scores for in-distribution and out-of-distribution tasks for each model. Detailed metric values for each task are provided in Appendix \ref{sec:appendix-sec-model-size}. As shown, larger backbone models generally exhibit better generalizability when fine-tuned across various tasks. Although cross-task fine-tuning can be employed across a variety of backbone models, a balance between hosting/inference cost and inference quality should be taken into account for business scenarios.

%% file: src/sec6-conclusion.tex
\vspace{-0.2cm}
\section{Conclusions}
\vspace{-0.2cm}
In this study, we introduced a cross-task fine-tuning approach and demonstrated its efficacy using publicly available content moderation datasets. Our findings reveal that fine-tuning a SLM exclusively with public content moderation data can yield robust performance in bespoke scenarios governed by custom guidelines. Furthermore, our approach enables knowledge transfer across tasks, even when the tasks are not closely related. By employing cross-task fine-tuning, we successfully developed a high-quality model that is comparable to GPT-3.5-Turbo on various tasks, and achieves nearly equivalent results to GPT-4-Turbo on brand new English binary classification tasks. This underscores the potential of our method as a competitive alternative in the realm of advanced language models.

%% file: src/sec8-appendix.tex
\section{Categories of Content Moderation Tasks}
\label{sec:categories}

Based on ~\citet{wang2023not}, we classified content moderation tasks into the following categories:
\vspace{-0.5cm}
\paragraph{Malicious Actions} This category encompasses tasks involve the modification of content that promotes or aids actions with potential harmful consequences. It can be divided into two subcategories: 1) Illegal Activities, which consist of content endorsing violence, threats, substance abuse, and more. 2) Unethical or Unsafe Actions, which cover content that encourages unhealthy practices, provides guidance for unsafe behaviors, or promotes harassment. Additionally, the text specifies that content pertaining to "Jailbreak" falls within this category, which includes attempts to circumvent safeguards and elicit unauthorized outputs from large language models.
\vspace{-0.15cm}
\paragraph{Discrimination, Exclusion, Toxicity, Hateful, Offensive} This category involves tasks for addressing harmful and toxic online content. There are three subcategories: 1) Social Stereotype and Unfair Discrimination, which involves content that propagates prejudices or stereotypes against specific groups of people. 2) Toxic Language (Hate Speech), including toxic or offensive language. 3) Adult Content, which consists of explicit sexual material and graphic violence.
\vspace{-0.15cm}
\paragraph{Information Hazards} Tasks which aims to identifying content that presents potential information risks falls under this category. For instance, it targets content that probes for personal, private details such as sensitive financial data or personal contact details, which could potentially lead to serious repercussions.
\vspace{-0.15cm}
\paragraph{Misinformation Harms} Tasks for detecting content that disseminates misleading information or causes potential harm belongs to this category. It consists of two subcategories: 1) Disseminating False Information, such as spam, rumors and propaganda. 2) Causing Material Harm through Misinformation, which involves content that shares unverified or uncorroborated information, potentially leading to medical or financial harm.

\section{Prompt Generation for Public Content Moderation Dataset}

\subsection{Definition generation}
\label{sec:definition-generation}

As outlined in Section~\ref{fine-tuning-models}, a guideline involves two components: the definitions of undesired content and the evaluation process. When the definitions are not readily available in the source dataset, we employ GPT-4-Turbo to generate these definitions.

This is accomplished through a few-shot learning process. For each sample, the prompt is the name of the undesired category (for instance, hate speech or violence), and the generated output is the definition of that category. We utilize the definitions from our private dataset as few-shot samples. The configuration of GPT-4-Turbo follows the setting mentioned in Appendix~\ref{sec:baseline-models}.

\subsection{Example prompts}
\label{sec:example-prompt}
The text below serves as a sample prompt for the \textit{Anatomy of Online Hate} dataset\cite{salminen2018anatomy}, aimed at classifying the targets of hate speech. The first part describes the task and provides definitions for each target as outlined in ~\citet{salminen2018anatomy}. The second part, starting with "Key Points to Consider", clarifies the evaluation process and the data format. The third part is the content to be reviewed, namely "<content>". 

\begin{quote}

\textbf{Hate Speech Target Classification Guidelines}
   
Your goal is to classify content based on its target, according to the 13 targets defined below. Only one target should be labeled for each content.  
   
\textbf{Definitions of Targets}

\begin{itemize}
    \item \textbf{None (Label: 0).} The content is not hateful, and therefore there is no target.
    \item \textbf{Financial Powers (Label: 1).} Hatred toward wealthy people and companies and their privileges. Pointing out their intentions to manipulate and commit crimes.
    \item \textbf{Political Issues (Label: 2).} Hate toward government, political parties and movements, war, terrorism, the flaws of the system.
    \item ...
\end{itemize}

\textbf{Key Points to Consider}
   
1. \textbf{Understanding the Categories:} Familiarize yourself with the definitions of the 13 harm categories. Understand what each category represents and the kind of content it covers.  
   
2. \textbf{Single Category Labeling:} Even if a content can be associated with multiple categories, only one category should be labeled for each content. Choose the most prominent or harmful category.  

3. \textbf{Input Format:} The content to be classified is located after "Content". This is the specific content that you should focus on for classification purposes.  
   
4. \textbf{Output Format:} Begin your classification response with the term "Label: ", followed by the index number of the category. Index numbering is indicated in the parentheses and starts from 0. For instance, if the content is classified under the category "None (Label: 0)", your response should be "Label: 0".

\textbf{Content}

<content>  
   
\textbf{Evaluation}

\end{quote}

\section{Implementation Details}
\label{sec:implementation-details}

\paragraph{Fine-tuning}
The fine-tuning process was conducted for 1 epoch, employing a batch size of 96. The learning rate is 1e-4 with a warm-up ratio of 0.03. For QLoRA, we set the rank to 64 and the scaling factor to 16. The dropout probability for LoRA is 0.05.

We chose Mistral-7B \cite{jiang2023mistral} v0.1, a 7-billion-parameter language model that has been open-sourced by Mistral AI, as the backbone model for STAND-Guard and evaluate its effectiveness compared to the vanilla model and some other baselines. We further evaluate the influence of the backbone model's size by comparing models cross-task fine-tuned on Phi-3-mini-128k-instruct, which has 3.8 billion parameters \cite{abdin2024phi}, and Mixtral-8×7 version 0.1 \cite{jiang2024mixtral}, with our STAND-Guard model. All the models adhere to identical training protocols mentioned above.

\paragraph{Inference}
During inference, we assign a top\_p value of 1.0, a temperature of 0.0 and a max\_tokens of 100 for all the models.

\paragraph{Metrics}

The \textbf{F1 score} is used as the evaluation metrics. For multi-class classification, we calculate the F1 metrics for each label, and find their average weighted by support (the number of true instances for each label)\footnote{https://scikit-learn.org/stable/modules/generated/\\sklearn.metrics.f1\_score.html}.

It should be noted that we classify any predictions that do not adhere to the schema outlined in the guideline as incorrect. Consequently, the F1 score are calculated based on the entire set of cases, rather than solely on those successfully parsed. Higher F1 values indicate better performance.

\section{Baseline Models}
\label{sec:baseline-models}
\vspace{-0.1cm}
\subsection{Task-specific models}
\vspace{-0.1cm}
These models are classifiers that are trained for specific tasks but do not accommodate the input of custom policies.
\vspace{-0.1cm}
\paragraph{Perspective API}\footnote{https://www.perspectiveapi.com/} offers services for the detection of toxic and hateful content. It encompasses a range of categories, such as toxicity, severe toxicity, insult, profanity, identity attacks, threats, and sexually explicit material. For the purpose of comparison, we convert the scores returned by the API into binary outcomes using a threshold of 0.5.
\vspace{-0.1cm}
\paragraph{OpenAI Content Moderation API}\cite{markov2023holistic} is trained to detect a set of categories of undesired content, including sexual content, hateful content, violence, self-harm, and harassment. Similar to the settings for Perspective API, we binarize the scores provided by the API with a threshold value of 0.5.
\vspace{-0.1cm}
\subsection{General models}
\vspace{-0.1cm}
These models refer to LLMs and SLMs that are designed to accept the guideline as an in-context input to steer the classification of the input text.
\vspace{-0.1cm}
\paragraph{LlamaGuard}\cite{inan2023llama} is fine-tuned for content moderation based on Llama2-7B. The first token of the output is adjusted to indicate if the content is "safe" or "unsafe", and the second token indicates the specific harmful category. We made slight modifications to the prompt's output schema for LlamaGuard to ensure compatibility with its pre-trained counterparts. 
\vspace{-0.1cm}
\paragraph{GPT-3.5-Turbo and GPT-4-Turbo}\footnote{https://platform.openai.com/docs/models/gpt-3-5-turbo, https://platform.openai.com/docs/models/gpt-4-turbo-and-gpt-4}. GPT-4 is considered to be the most powerful LLM to date, and GPT-4-Turbo is a new version that supports longer context. There is ongoing work to integrate GPT models for content moderation\footnote{https://openai.com/index/using-gpt-4-for-content-moderation/}. For both GPT-3.5-Turbo and GPT-4-Turbo, in addition to the common configuration shared by all models, we configured frequency\_penalty and presence\_penalty to 0. 


\section{Task Analysis}
\label{sec:dataset-correlation}

Content moderation tasks are classified based on categories and subcategories described in Section~\ref{sec:methodology-building-training-set}. Table~\ref{tab:data-analysis-1} and~\ref{tab:data-analysis-2} show a detailed analysis of all the tasks used in this study. A mark (\cmark) denotes that the undesired content overlaps with the subcategory. It is noteworthy that even though some tasks share the same name, the definition of the undesired content can be different, highlighting the importance of developing a model that quickly adapts to diverse content moderation tasks.

\begin{sidewaystable*} \tiny
\begin{tabular}{ll|cc|ccc|c|cc}
\toprule
 &  & \multicolumn{2}{c}{\textbf{Malicious Actions}} & \multicolumn{3}{c}{\textbf{Discrimination,   Exclusion, Toxicity, Hateful, Offensive}} & \textbf{Information Hazards} & \multicolumn{2}{c}{\textbf{Misinformation Harms}} \\
 &  & {\textbf{\begin{tabular}[c]{@{}l@{}}Illegal \\ \end{tabular}}} & {\textbf{Unethical or}} & {\textbf{Social Stereotype}} & {\textbf{Toxic Language}} & {\textbf{Adult Content}} & {\textbf{Private}} & {\textbf{Disseminate False}} & {\textbf{Causing Material Harm}} \\
 &  & {\textbf{Activities}} & {\textbf{Unsafe Actions}} & {\textbf{and Unfair}} & {\textbf{(Hate Speech)}} & {\textbf{}} & {\textbf{Information}} & {\textbf{or Misleading}} & {\textbf{by Disseminating}} \\
\multirow{-4}{*}{\textbf{Dataset}} & \multirow{-4}{*}{\textbf{Task}} & {\textbf{}} & {\textbf{}} & {\textbf{Discrimination}} & {\textbf{}} & {\textbf{}} & {\textbf{}} & {\textbf{Information}} & {\textbf{Unverified Information}} \\
\midrule
Anatomy of Online Hate & Hate &  &  &  & \cmark &  &  &  &  \\
BIG-bench (German) & Gender Inclusive &  &  & \cmark &  &  &  &  &  \\
BIG-bench (Hinglish) & Toxic &  &  &  & \cmark &  &  &  &  \\
CallMeSexist & Sexism &  &  & \cmark &  &  &  &  &  \\
Civil-Comments & Insult &  &  &  & \cmark &  &  &  &  \\
 & Obscenity &  &  &  & \cmark &  &  &  &  \\
 & Severe Toxicity &  &  &  & \cmark &  &  &  &  \\
 & Sexually Explicit &  &  &  &  & \cmark &  &  &  \\
 & Threat & \cmark &  &  &  &  &  &  &  \\
 & Toxicity &  &  &  & \cmark &  &  &  &  \\
Commonsense Morality & Ethics &  & \cmark &  &  &  &  &  &  \\
COVID-HATE & Hate &  &  &  & \cmark &  &  & \cmark &  \\
CrowS-Pairs & Bias &  &  & \cmark &  &  &  &  &  \\
DecodingTrust & Stereotype &  &  & \cmark &  &  &  &  &  \\
DynaHate & Hate &  &  &  & \cmark &  &  &  &  \\
Exaggerated Safety & Safety & \cmark & \cmark & \cmark &  &  & \cmark &  &  \\
GermEval (German) & Offensive &  &  &  & \cmark &  &  &  &  \\
HASOC (English) & Hate, Offensive &  &  &  & \cmark &  &  &  &  \\
HASOC (German) & Hate, Offensive &  &  &  & \cmark &  &  &  &  \\
Hate Speech and Offensive Language & Hate, Offensive &  &  &  & \cmark &  &  &  &  \\
Hate Speech towards Foreigners (German) & Hate &  &  &  & \cmark &  &  &  &  \\
HateCheck & Hate &  &  &  & \cmark &  &  &  &  \\
HateEval & Hate &  &  &  & \cmark &  &  &  &  \\
HatemojiCheck & Hate &  &  &  & \cmark &  &  &  &  \\
HateXplain & Hate & \cmark &  &  & \cmark &  &  &  &  \\
Jigsaw & Identity Hate &  &  & \cmark & \cmark &  &  &  &  \\
 & Insult &  &  &  & \cmark &  &  &  &  \\
 & Obscene &  &  &  & \cmark &  &  &  &  \\
 & Severe Toxic &  &  &  & \cmark &  &  &  &  \\
 & Threat & \cmark &  &  &  &  &  &  &  \\
 & Toxic &  &  &  & \cmark &  &  &  &  \\
Jiminy Cricket & Ethics &  & \cmark &  &  &  &  &  &  \\
Korean Hate Speech (Korean) & Hate &  &  & \cmark & \cmark &  &  &  &  \\
 & Aggressiveness &  &  &   & \cmark &  &  &  &  \\
& Bias &  &  & \cmark &  &  &  &  &  \\
OffComBR3 (Portuguese) & Offensive &  &  & \cmark & \cmark &  &  &  &  \\
OpenAI CM & Harassment &  & \cmark &  &  &  &  &  &  \\
 & Hateful & \cmark &  &  & \cmark &  &  &  &  \\
 & Self Harm &  & \cmark &  &   &  &  &  &  \\
 & Sexual &  &  &  &  & \cmark &  &  &  \\
 & Violence & \cmark &  &  &  & \cmark &  &  &  \\
\bottomrule
\end{tabular}
\caption{\label{tab:data-analysis-1}Task analysis based on the task taxonomy. A mark (\cmark) denotes that the definition of the undesired content overlaps with the subcategory of the content moderation task.}
\end{sidewaystable*}

\begin{sidewaystable*} \tiny
\begin{tabular}{ll|cc|ccc|c|cc}
\toprule
 &  & \multicolumn{2}{c}{\textbf{Malicious Actions}} & \multicolumn{3}{c}{\textbf{Discrimination, Exclusion, Toxicity, Hateful, Offensive}} & \textbf{Information Hazards} & \multicolumn{2}{c}{\textbf{Misinformation Harms}} \\
 &  & {\textbf{\begin{tabular}[c]{@{}l@{}}Illegal \\ \end{tabular}}} & {\textbf{Unethical or}} & {\textbf{Social Stereotype}} & {\textbf{Toxic Language}} & {\textbf{Adult Content}} & {\textbf{Private}} & {\textbf{Disseminate False}} & {\textbf{Causing Material Harm}} \\
 &  & {\textbf{Activities}} & {\textbf{Unsafe Actions}} & {\textbf{and Unfair}} & {\textbf{(Hate Speech)}} & {\textbf{}} & {\textbf{Information}} & {\textbf{or Misleading}} & {\textbf{by Disseminating}} \\
\multirow{-4}{*}{\textbf{Dataset}} & \multirow{-4}{*}{\textbf{Task}} & {\textbf{}} & {\textbf{}} & {\textbf{Discrimination}} & {\textbf{}} & {\textbf{}} & {\textbf{}} & {\textbf{Information}} & {\textbf{Unverified Information}} \\
\midrule
PKU-Alignment Safe-RLHF & Unsafe & \cmark & \cmark & \cmark & \cmark & \cmark & \cmark & \cmark & \cmark \\
PKU-Alignment BeaverTails & Animal Abuse & \cmark &  &  &  &  &  &  &  \\
 & Child Abuse & \cmark &  &  &  &  &  &  &  \\
 & Controversial Topics, Politics &  &  &  &  &  &  & \cmark &  \\
 & Discrimination, Stereotype &  &  & \cmark &  &  &  &  &  \\
 & Drug Abuse, Weapons & \cmark &  &  &  &  &  &  &  \\
 & Financial \& Property Crime & \cmark &  &  &  &  &  &  &  \\
 & Hateful \& Offensive   Language &  &  &   & \cmark &  &  &  &  \\
 & Misinformation & \cmark & \cmark &   &  &  &  & \cmark & \cmark \\
 & Non-Violent Unethical Behavior &  & \cmark &  &  &  &  &  &  \\
 & Privacy Violation &  &  &  &  &  & \cmark &  &  \\
 & Self Harm &  & \cmark &  &  &  &  &  &  \\
 & Sexually Explicit &  &  &  &  & \cmark &  &  &  \\
 & Terrorism, Organized Crime & \cmark &  &  &  &  &  &  &  \\
 & Violence & \cmark &  &  &  &  &  &  &  \\
PKU-Alignment-BeaverTails-Eval & Unsafe & \cmark & \cmark & \cmark & \cmark & \cmark & \cmark & \cmark & \cmark \\
Private & Hate &  &  &  & \cmark &  &  &  &  \\
 & Self Harm &  & \cmark &  &  &  &  &  &  \\
 & Sexual &  &  &  &  & \cmark &  &  &  \\
 & Violence & \cmark &  &  &  &  &  &  &  \\
 & Jailbreak & & \cmark &  &  &  &  &  & \\
Reddit Content Moderation & Rule Moderation & \cmark & \cmark & \cmark & \cmark & \cmark & \cmark & \cmark &  \\
RP-Mod \& RP-Crowd (German) & Offensive & \cmark &  & \cmark & \cmark &  &  & \cmark &  \\
Scruples Anecdotes & Ethics &  & \cmark &  &  &  &  &  &  \\
Social Bias Inference Corpus (SBIC) & Intentionally Offensive &  &  &  & \cmark &  &  &  &  \\
 & Potentially Offensive &  &  &  & \cmark &  &  &  &  \\
 & Sexually Offensive &  &  &  & \cmark &  &  &  &  \\
SWAD & Swear &  &  &  & \cmark &  &  &  &  \\
SWSR (Chinese) & Sexism &  &  & \cmark &  &  &  &  &  \\
ToLD-BR (Portuguese) & Offensive &  &  & \cmark & \cmark &  &  &  &  \\
 & Homophobia &  &  & \cmark &  &  &  &  &  \\
 & Misogyny &  &  & \cmark &  &  &  &  &  \\
 & Racism &  &  & \cmark &  &  &  &  &  \\
ToxiGen & Toxic &  &  &  & \cmark &  &  &  &  \\
TrustworthyLLM & Safety & \cmark & \cmark &  &  & \cmark & \cmark &  &  \\
TweetEval & Hate &  &  &  & \cmark &  &  &  &  \\
 & Irony &  & \cmark &  &  &  &  &  &  \\
 & Offensive &  &  &  & \cmark &  &  &  &  \\
USElectionHate & Hate & \cmark &  & \cmark & \cmark &  &  &  &  \\
White Supremacist & Hate &  &  & \cmark & \cmark &  &  &  &  \\
\bottomrule
\end{tabular}
\caption{\label{tab:data-analysis-2}Task analysis based on the task taxonomy (continued). A mark (\cmark) denotes that the definition of the undesired content overlaps with the subcategory of the content moderation task.}
\end{sidewaystable*}

\section{Data Statistics}

\subsection{Training Data}

The statistics of training data is presented in Table \ref{tab:train-dataset-stats}. We conducted strategic sampling to guarantee that each task is represented in roughly equal proportions within the full training dataset. 

\begin{table*}\scriptsize
  \centering
\begin{tabular}{llllrrr}
\toprule
Dataset & \#Task & Task & Binary/Multiple Class(es) & \#Data & \#Data (\%) & Harmful Ratio \\
\midrule
PKU-Alignment BeaverTails & 14     & Animal Abuse & binary  & 10,000 & 5.7\%  & 1.3\%   \\
&   & Child Abuse  & binary  & 9,949  & 5.7\%  & 0.7\%   \\
&   & Controversial Topics, Politics & binary  & 9,981  & 5.7\%  & 3.5\%   \\
&   & Discrimination, Stereotype     & binary  & 9,984  & 5.7\%  & 8.8\%   \\
&   & Drug Abuse, Weapons  & binary  & 10,000 & 5.7\%  & 5.6\%   \\
&   & Financial \& Property Crime    & binary  & 9,942  & 5.7\%  & 9.2\%   \\
&   & Hateful \& Offensive Language  & binary  & 10,000 & 5.7\%  & 9.2\%   \\
&   & Misinformation    & binary  & 10,000 & 5.7\%  & 2.3\%   \\
&   & Non-Violent Unethical Behavior & binary  & 9,964  & 5.7\%  & 17.4\%  \\
&   & Privacy Violation & binary  & 9,981  & 5.7\%  & 5.4\%   \\
&   & Self Harm    & binary  & 9,943  & 5.7\%  & 0.8\%   \\
&   & Sexually Explicit & binary  & 9,941  & 5.7\%  & 2.5\%   \\
&   & Terrorism, Organized Crime     & binary  & 10,000 & 5.7\%  & 1.1\%   \\
&   & Violence     & binary  & 9,976  & 5.7\%  & 25.5\%  \\
PKU-Alignment SafeRLHF& 1 & Unsafe  & binary  & 10,000 & 5.7\%  & 43.0\%  \\
Private & 5 & Hate & binary  & 3,733  & 2.1\%  & 19.1\%  \\
&   & Self harm    & binary  & 3,739  & 2.1\%  & 11.9\%  \\
&   & Sexual  & binary  & 3,794  & 2.2\%  & 17.9\%  \\
&   & Violence     & binary  & 3,734  & 2.1\%  & 18.1\%  \\
&   & Jailbreak    & binary  & 10,000 & 5.7\%  & 10.4\% 
\\\bottomrule
\end{tabular}
\vspace{-0.2cm}
  \caption{Statistics of training data. \#Data and \#Data (\%) indicate the number and proportion of training samples for the task, respectively. The Harmful Ratio indicates the proportion of entries with positive labels for the task.}
  \label{tab:train-dataset-stats}
\end{table*}

\subsection{Testing Data}

Table \ref{tab:test-dataset-stats} shows the statistics of tasks in the test set. 

\label{sec:stats-of-test-tasks}
\begin{table*}\tiny
  \centering
\begin{tabular}{llllrr}
\toprule
Dataset & \#Task & Task & Binary/Multiple Class(es) & \#Data & Harmful Ratio \\
\midrule
PKU-Alignment BeaverTails & 14     & Animal Abuse & binary  & 3,021  & 1.5\%  \\
      &  & Child Abuse  & binary  & 3,021  & 0.9\%  \\
      &  & Controversial Topics, Politics & binary  & 3,021  & 3.1\%  \\
      &  & Discrimination, Stereotype     & binary  & 3,021  & 9.8\%  \\
      &  & Drug Abuse, Weapons    & binary  & 3,021  & 5.0\%  \\
      &  & Financial \& Property Crime    & binary  & 3,021  & 8.7\%  \\
      &  & Hateful \& Offensive Language  & binary  & 3,021  & 10.0\% \\
      &  & Misinformation   & binary  & 3,021  & 2.5\%  \\
      &  & Non-Violent Unethical Behavior & binary  & 3,021  & 20.1\% \\
      &  & Privacy Violation      & binary  & 3,021  & 5.1\%  \\
      &  & Self Harm    & binary  & 3,021  & 0.6\%  \\
      &  & Sexually Explicit      & binary  & 3,021  & 3.4\%  \\
      &  & Terrorism, Organized Crime     & binary  & 3,021  & 1.4\%  \\
      &  & Violence     & binary  & 3,021  & 24.0\% \\
PKU-Alignment Safe-RLHF  & 1      & Unsafe & binary  & 66,088 & 53.5\% \\
Private & 4      & Hate & binary  & 1,000  & 30.3\% \\
      &  & Self harm    & binary  & 1,000  & 19.7\% \\
      &  & Sexual & binary  & 1,000  & 19.2\% \\
      &  & Violence     & binary  & 1,000  & 26.3\% \\
OpenAI CM     & 5      & Harassment   & binary  & 1,444  & 5.3\%  \\
      &  & Hateful      & multiple  & 1,680  & 23.9\% \\
      &  & Self Harm    & binary  & 1,447  & 3.5\%  \\
      &  & Sexual & multiple  & 1,680  & 25.7\% \\
      &  & Violence     & multiple  & 1,680  & 6.2\%  \\
TweetEval     & 3      & Hate & binary  & 2,970  & 42.2\% \\
      &  & Irony  & binary  & 784    & 39.7\% \\
      &  & Offensive    & binary  & 860    & 27.9\% \\
Jigsaw  & 5      & Identity Hate  & binary  & 63,978 & 1.1\%  \\
      &  & Toxic  & binary  & 63,978 & 9.5\%  \\
      &  & Threat & binary  & 63,978 & 0.3\%  \\
      &  & Insult & binary  & 63,978 & 5.4\%  \\
      &  & Obscenity    & binary  & 63,978 & 5.8\%  \\
      &  & Severe Toxicity  & binary  & 63,978 & 0.6\%  \\
White Supremacist \cite{gibert2018hate}   & 1      & Hate & binary  & 478    & 50.0\% \\
Anatomy of Online Hate \cite{salminen2018anatomy} & 1      & Hate & multiple  & 3,222  & 73.4\% \\
BIG-bench (German) \cite{srivastava2023beyond}     & 1      & Gender Inclusive & binary  & 489    & 40.9\% \\
CallMeSexist \cite{samory2021call}   & 1      & Sexism & binary  & 13,631 & 13.3\% \\
Civil-Comments \cite{DBLP:journals/corr/abs-1903-04561}   & 6      & Insult & binary  & 2,997  & 49.9\% \\
      &  & Obscenity    & binary  & 1,998  & 49.9\% \\
      &  & Severe Toxicity  & binary  & 2,985  & 49.9\% \\
      &  & Sexually Explicit      & binary  & 1,990  & 50.2\% \\
      &  & Threat & binary  & 1,996  & 50.1\% \\
      &  & Toxicity     & binary  & 2,997  & 49.9\% \\
Commonsense Morality \cite{hendrycks2021aligning}  & 1      & Ethics & binary  & 3,885  & 46.7\% \\
COVID-HATE \cite{he2021racism}     & 1      & Hate & multiple  & 2,290  & 41.3\% \\
CrowS-Pairs \cite{nangia2020crows}    & 1      & Bias & binary  & 1,508  & 100.0\%      \\
DecodingTrust \cite{wang2023decodingtrust}  & 1      & Stereotype   & binary  & 1,152  & 100.0\%      \\
DynaHate \cite{vidgen2021learning} & 1      & Hate & binary  & 41,255 & 54.0\% \\
Exaggerated Safety \cite{rottger2023xstest}      & 1      & Safety & binary  & 450    & 44.4\% \\
GermEval (German) bin \cite{wiegand2018overview}  & 1      & Offensive    & binary  & 3,532  & 34.0\% \\
GermEval (German) multi \cite{wiegand2018overview}  & 1      & Offensive    & multiple  & 3,532  & 89.2\% \\
HASOC   (English) \cite{mandl2019overview} & 1      & Hate, Offensive  & binary  & 1,153  & 25.0\% \\
HASOC   (German) \cite{mandl2019overview}     & 1      & Hate, Offensive  & binary  & 3,819  & 10.7\% \\
Hate Speech and Offensive   Language     & 1      & Hate, Offensive  & multiple  & 24,783 & 94.2\% \\
\cite{davidson2017automated} &&&&\\
Hate Speech towards Foreigners (German) & 1      & Hate & multiple  & 666    & 100.0\%      \\
\cite{bretschneider2017detecting}  &&&& \\
HateCheck \cite{rottger2020hatecheck}    & 1      & Hate & binary  & 3,901  & 68.2\% \\
HateEval \cite{basile2019semeval}     & 1      & Hate & binary  & 4,571  & 41.8\% \\
HatemojiCheck \cite{kirk2021hatemoji}  & 1      & Hate & binary  & 3,930  & 67.5\% \\
HateXplain \cite{mathew2021hatexplain}     & 1      & Hate & binary  & 1,924  & 59.4\% \\
Jiminy-Cricket \cite{hendrycks2021would}   & 1      & Ethics & multiple  & 3,986  & 50.4\% \\
Korean   Hate Speech (Korean) \cite{moon2020beep}    & 3      & Hate & binary  & 471    & 68.4\% \\
      &  & Aggressiveness  & multiple  & 471    & 66.0\% \\
      &  & Bias & multiple  & 471    & 27.4\% \\
OffComBR3 (Portuguese) \cite{Pelle2017} & 1      & Offensive    & binary  & 1,250  & 33.5\% \\
PKU-Alignment-BeaverTails-Eval \cite{ji2024beavertails}   & 1      & Unsafe & multiple  & 700    & 92.9\% \\
Reddit Content Moderation \cite{kumar2024watch}      & 1      & Rule Moderation  & binary  & 96,544 & 50.2\% \\
RP-Mod \& RP-Crowd (German)    & 1      & Offensive    & binary  & 57,410 & 50.0\% \\
 \cite{Assenmacher2021RPModRPCrowdM} &&&&\\
Scruples Anecdotes \cite{Lourie2020Scruples}     & 1      & Ethics & binary  & 6,159  & 25.0\% \\
Social   Bias Inference Corpus (SBIC) & 3      & Intentionally Offensive  & binary  & 3,462  & 50.0\% \\
    \cite{sap2019social} &  & Potentially Offensive  & binary  & 5,892  & 50.0\% \\
      &  & Sexually Offensive     & binary  & 3,462  & 50.0\% \\
SWAD \cite{pamungkas2020you} & 1      & Swear  & binary  & 2,577  & 32.7\% \\
SWSR (Chinese) bin \cite{jiang2022swsr}      & 1      & Sexism & binary  & 8,969  & 34.5\% \\
SWSR (Chinese) multi \cite{jiang2022swsr}   & 1      & Sexism & multiple  & 8,969  & 34.5\% \\
ToLD-BR (Portuguese) \cite{leite2020toxic}    & 4      & Offensive    & binary  & 21,000 & 44.1\% \\
      &  & Homophobia   & binary  & 21,000 & 1.6\%  \\
      &  & Misogyny     & binary  & 21,000 & 2.2\%  \\
      &  & Racism & binary  & 21,000 & 0.7\%  \\
ToxiGen \cite{hartvigsen2022toxigen} & 1      & Toxic  & binary  & 940    & 43.2\% \\
TrustworthyLLM \cite{liu2023trustworthy}  & 1      & Safety & binary  & 5,904  & 14.0\% \\
USElectionHate\cite{USElectionHate}   & 1      & Hate & binary  & 600    & 9.8\% 
\\\bottomrule
\end{tabular}
  \caption{Statistics of test data. \#Data and \#Data (\%) indicate the number and proportion of training samples for the task, respectively. For binary classification tasks, the Harmful Ratio indicates the proportion of entries with positive labels for the task. For multi-class tasks, the Harmful Ratio takes into account all instances that are marked with positive labels.}
  \label{tab:test-dataset-stats}
\end{table*}

\section{Error Analysis of GPT-4}
\label{sec:gpt4-error}

The performance difference on in-distribution tasks between zero-shot GPT-4-Turbo and STAND-guard is considerable (0.68 vs 0.46). Given that GPT-4-Turbo possesses outstanding zero-shot capability, it is necessary to delve into the errors of GPT-4-Turbo.

A common error in the PKU-Alignment Safe-RLHF dataset is role misinterpretation. This dataset features conversations between a user and a bot, with only the bot’s response being subject to modification. GPT-4-Turbo often misclassifies based on the user’s input rather than the bot’s response. In contrast, STAND-Guard, having encountered such conversations during training, accurately identifies and flags inappropriate bot responses.

Another error is target generalization, which occurs when guidelines focus on a specific target (e.g., child abuse), but the content is more general (e.g., abuse). GPT-4-Turbo tends to incorrectly flag such content. However, STAND-Guard, due to its fine-tuning process, is better equipped to handle these nuances.

\section{Comparison with Task-Specific Baselines}
\label{sec:comparison-task-specific}
\begin{table*}[]\scriptsize
\centering
\begin{tabular}{llrrrrrr|r}
\toprule
Dataset &Task &Perspective &OpenAI CM &LlamaGuard &GPT-3.5-Turbo &GPT-4-Turbo &Mistral-7B &\textbf{STAND-Guard} \\
\midrule
Jigsaw (sampled) &Identity Hate &0.278 &0.579 &0.214 &0.098 &0.236 &0.018 &0.255 \\
 &Insult &0.482 &0.498 &0.469 &0.233 &0.425 &0.077 &0.410 \\
 &Obscene &0.531 &0.225 &0.161 &0.304 &0.545 &0.081 &0.508 \\
 &Threat &0.159 &0.359 &0.243 &0.041 &0.281 &0.004 &0.302 \\
 &Toxic &0.119 &0.584 &0.529 &0.364 &0.451 &0.127 &0.558 \\
OpenAI CM &Harassment &0.290 &0.327 &0.161 &0.255 &0.268 &0.077 &0.726 \\
 &Hateful &0.716 &0.732 &0.729 &0.669 &0.671 &0.577 &0.732 \\
 &Self Harm &- &0.891 &0.000 &0.292 &0.630 &0.099 &0.928 \\
 &Sexual &0.655 &0.755 &0.742 &0.696 &0.742 &0.461 &0.475 \\
 &Violence &0.922 &0.949 &0.927 &0.815 &0.707 &0.674 &0.671 \\
TweetEval &Hate &0.249 &0.295 &0.650 &0.686 &0.486 &0.308 &0.556 \\
 &Irony &- &- &0.061 &0.685 &0.780 &0.005 &0.685 \\
 &Offensive &0.614 &0.480 &0.381 &0.582 &0.525 &0.573 &0.682 \\
White Supremacist &Hate &0.584 &0.508 &0.503 &0.796 &0.711 &0.582 &0.739 \\
\hline
\multicolumn{2}{c}{AVG} &0.467 &0.552 &0.412 &0.465 &0.533 &0.262 & 0.588
 \\\bottomrule
\end{tabular}
    \vspace{-0.2cm}
\caption{\label{tab:out-of-distribution-results-hate}
    F1 scores on out-of-distribution tasks related to hate speech and offensive language detection under zero-shot setting. Task-specific baselines (Perspective and OpenAI CM) are included in the comparison.
  }
\end{table*}

Table \ref{tab:out-of-distribution-results-hate} contains the results obtained using task-specific API models on datasets concerning hate speech and offensive language, in accordance with the methodology outlined by \citet{markov2023holistic}. 

\section{Case Study}
\label{sec:exp-casestudy}
\vspace{-0.15cm}

Cross-task fine-tuning enables the model to understand derogatory implications. The significant classification gain on the dataset \textit{CallMeSexist}, which bears scant resemblance to the training set, serves as a case in point. Consider the following statement from  \textit{CallMeSexist}: \textit{For real. Asking a woman to move smh!! naw no go"MENTION3988: MENTION3549 lmao IDC IDC! Call your male friends."} It is not explicitly demeaning towards women. Instead, it subtly suggests that a woman's presence is less important or valued than a man's. STAND-Guard succeeds in detecting implicit harmful intent in the statement.

Without external knowledge, all models struggle to flag the undesired content. For instance, when detecting controversial or political topics in \textit{PKU-Alignment BeaverTails}, even GPT-4-Turbo achieves a F1 of only 0.043. This could be attributed to the intrinsic nature of the task. The conversations in this task encompass political figures, policies and current affairs, such as \textit{Josef Mengele is most known for his horrific human experiments at Auschwitz during WWII, ...}. To make correct predictions, knowledge about these topics are necessary, posing challenges for pre-trained models. 

\section{Multilingual Tasks}
\label{sec:multilingual}

Tables \ref{tab:out-of-distribution-results-non-en} shows the experimental results on non-English, out-of-distribution tasks under zero-shot setting.

\begin{table*}[]\scriptsize
  \centering
\begin{tabular}{llrrrr|r}
\toprule
{Dataset} &{Task} &{LlamaGuard} &{GPT-3.5-Turbo} &{GPT-4-Turbo} &Mistral-7B &{\textbf{STAND-Guard}} \\
 \midrule
BIG-bench (German) &Gender Inclusive &0.000 &0.855 &0.851 &0.000 &0.708 \\
GermEval (German) bin &Offensive &0.012 &0.679 &0.670 &0.574 &0.501 \\
GermEval (German) multi &Offensive &0.060 &0.761 &0.734 &0.190 &0.101 \\
HASOC   (German) &Hate, Offensive &0.038 &0.219 &0.302 &0.217 &0.417 \\
Hate Speech towards Foreigners   (German) &Hate &0.279 &0.566 &0.674 &0.532 &0.495 \\
Korean   Hate Speech (Korean) &Hate &0.206 &0.712 &0.824 &0.012 &0.240 \\
 &Aggressiveness &0.300 &0.550 &0.705 &0.539 &0.382 \\
 &Bias &0.736 &0.706 &0.739 &0.458 &0.698 \\
OffComBR3 (Portuguese) &Offensive &0.340 &0.640 &0.722 &0.424 &0.558 \\
RP-Mod \& RP-Crowd (German) &Offensive &0.467 &0.301 &0.358 &0.496 &0.210 \\
SWSR (Chinese) bin &Sexism &0.035 &0.602 &0.568 &0.211 &0.504 \\
SWSR (Chinese) multi &Sexism &0.567 &0.576 &0.625 &0.182 &0.528 \\
ToLD-BR (Portuguese) &Offensive &0.459 &0.656 &0.693 &0.482 &0.302 \\
 &Homophobia &0.093 &0.130 &0.428 &0.034 &0.140 \\
 &Misogyny &0.004 &0.153 &0.297 &0.023 &0.624 \\
 &Racism &0.062 &0.055 &0.258 &0.015 &0.215 \\
 \hline
\multicolumn{2}{c}{AVG} &0.229 &0.510 &0.591 &0.274 &0.414
\\\bottomrule
\end{tabular}
\vspace{-0.2cm}
  \caption{\label{tab:out-of-distribution-results-non-en}
     F1 scores on out-of-distribution tasks (non-English data) under zero-shot setting.
  }
\end{table*}

\section{Multi-class Classification Tasks}
\label{sec:multi-class}

Tables \ref{tab:out-of-distribution-results-multiclass} shows the experimental results on multi-class classification, out-of-distribution tasks under zero-shot setting.

\begin{table*}[]\scriptsize
  \centering
\begin{tabular}{llrrrr|r}
\toprule
{Dataset} &{Task} &{LlamaGuard} &{GPT-3.5-Turbo} &{GPT-4-Turbo} &Mistral-7B &{\textbf{STAND-Guard}} \\
 \midrule
Anatomy of Online Hate &Hate &0.255 &0.278 &0.541 &0.183 &0.269 \\
COVID-HATE &Hate &0.434 &0.085 &0.796 &0.503 &0.695 \\
GermEval (German) multi &Offensive &0.060 &0.761 &0.734 &0.190 &0.101 \\
Hate Speech and Offensive   Language &Hate, Offensive &0.497 &0.799 &0.870 &0.655 &0.585 \\
Hate Speech towards Foreigners   (German) &Hate &0.279 &0.566 &0.674 &0.532 &0.495 \\
Jiminy-Cricket &Ethics &0.329 &0.633 &0.646 &0.407 &0.598 \\
Korean   Hate Speech (Korean) &Aggressiveness &0.300 &0.550 &0.705 &0.539 &0.382 \\
 &Bias &0.736 &0.706 &0.739 &0.458 &0.698 \\
OpenAI CM &Hateful &0.729 &0.669 &0.671 &0.577 &0.732 \\
 &Sexual &0.742 &0.696 &0.742 &0.461 &0.475 \\
 &Violence &0.927 &0.815 &0.707 &0.674 &0.671 \\
PKU-Alignment-BeaverTails-Eval &Unsafe &0.150 &0.467 &0.448 &0.136 &0.513 \\
SWSR (Chinese) multi &Sexism &0.567 &0.576 &0.625 &0.182 &0.528 \\
\hline
\multicolumn{2}{c}{AVG} &0.462 &0.585 &0.684 &0.423 &0.519
\\\bottomrule
\end{tabular}
\vspace{-0.2cm}
  \caption{\label{tab:out-of-distribution-results-multiclass}
    F1 scores on out-of-distribution tasks (multi-class classification) under zero-shot setting.
  }
\end{table*}

\section{Visualization of Task Semantic Similarity and Classification Gains}
\label{sec:vis-task-similarity}

\begin{figure}
    \centering
    \includegraphics[width=0.45\textwidth]{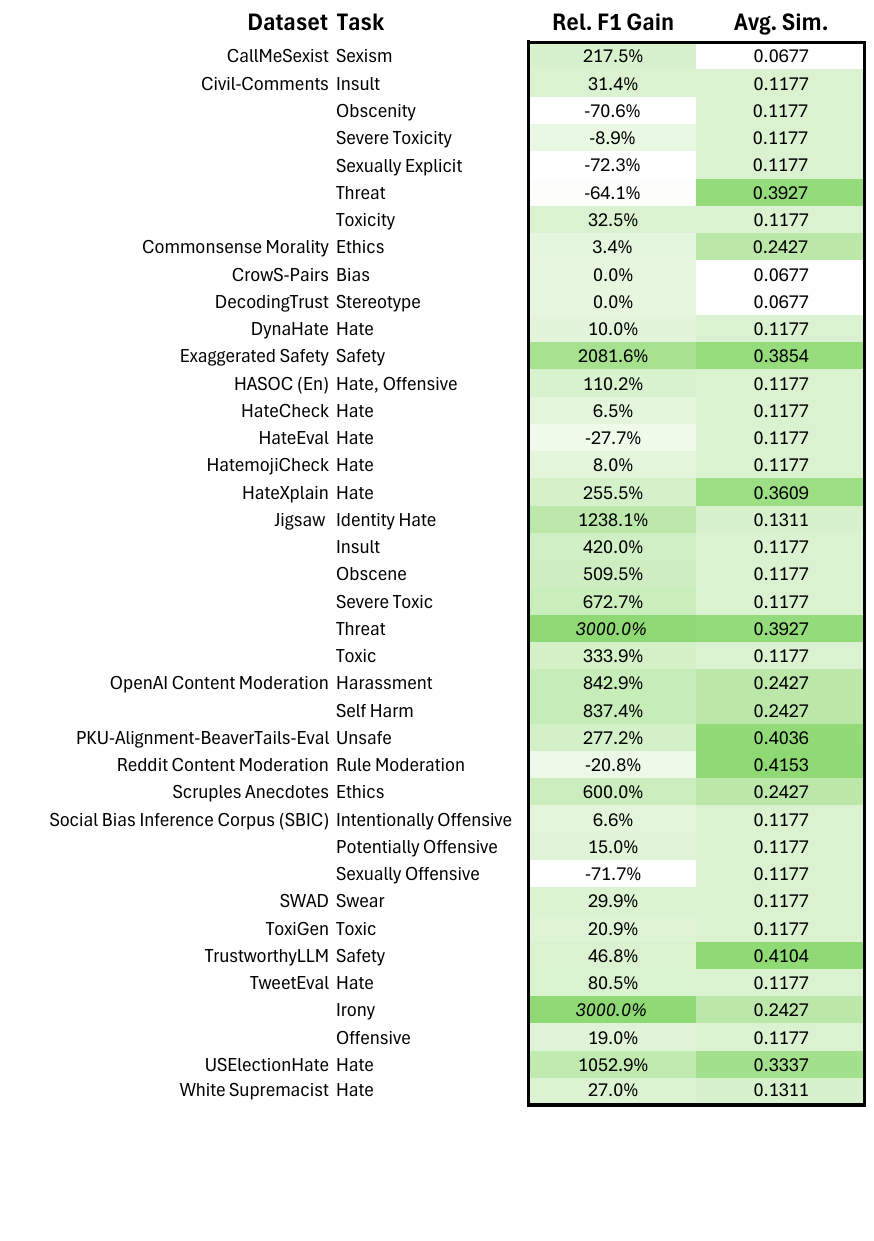}
    \caption{Heatmap between task semantic similarities and relative performance gains. The semantic similarities are calculated based on Table 9 and Table 10. For visualization, relative gains greater than 3000\% are set to 3000\% and marked in italic.}
    \label{fig:task-similarity}
\end{figure}

We further explore the correlation between task semantic similarity and classification quality improvements on English binary classification tasks in greater depth in Figure~\ref{fig:task-similarity}. We obtain these semantic similarities by representing each task as a binary vector, as per the data presented in Appendix~\ref{sec:dataset-correlation}. Following this, we calculate the cosine similarity between these tasks to determine their relative similarity. The analysis indicates that the fine-tuned model realized significant enhancements in classification quality over the vanilla model for tasks that closely resemble the training data, including \textit{Exaggerated Safety}, \textit{Jigsaw - Threat}, \textit{TweetEval - Irony}, and \textit{USElectionHate}. Intriguingly, we also observed benefits from cross-task knowledge transfer on test tasks that deviate from the training tasks in similarity, such as \textit{CallMeSexist} and \textit{Jigsaw}. This suggests that the fine-tuning process imparts a degree of generalizability to the model, allowing it to effectively adapt and perform well even on tasks that are not directly semantically aligned with the original training data.

\section{In-context Learning Capability}
\label{sec:in-context-learning}
\begin{table*}\scriptsize
\centering
\begin{tabular}{lllll|lll}
\toprule
\multirow{2}{*}{Dataset} & \multirow{2}{*}{Task} & \multicolumn{3}{c}{STAND-Guard w/ RAG} & \multicolumn{3}{c}{STAND-Guard} \\
     &     & F1      & Prec     & Rec     & F1   & Prec   & Rec   \\
\midrule
PKU-Alignment-BeaverTails-Eval & Unsafe  & 0.583 & 0.665 & 0.593 & 0.513 & 0.695 & 0.526 \\
Korean Hate Speech (Korean)  & Hate  & 0.784 & 0.913 & 0.686 & 0.240 & 0.978 & 0.137 \\
     & Aggressiveness & 0.941 & 0.944 & 0.941 & 0.382 & 0.484 & 0.450 \\
     & Bias  & 0.998 & 0.998 & 0.998 & 0.698 & 0.758 & 0.766 
\\\bottomrule
\end{tabular}
    \caption{F1 scores, precisions, recalls on out-of-distribution tasks under few-shot (w/ RAG) and zero-shot setting. The results show that fine-tuned task-adaptive model retains in-context learning ability.}
    \label{tab:exp-rag}
\end{table*}
We carried out further experiments to demonstrate that STAND-Guard, once fine-tuned via cross-task learning, retains the ability to perform in-context learning for new tasks. Utilizing the out-of-distribution datasets and tasks outlined in Section \ref{sec:dataset-stats}, we chose a subset of tasks for which training data was originally available (but not included in our training set) and applied Retrieval Augmented Generation (RAG) \cite{an2023context, hu2022context} to facilitate annotation for the SLM. Specifically, we dynamically selected 10 few-shot samples that are the most relevant to the current content requiring classification. This relevance was determined by calculating the cosine similarity between the training sample's embedding and the embedding of the content under review.

Table \ref{tab:exp-rag} presents a comparative analysis of the performance of the same model, fine-tuned through cross-task learning, with and without the incorporation of Retrieval Augmented Generation (RAG). As anticipated, the results confirm that STAND-Guard retains its in-context learning capabilities. Furthermore, the integration of RAG provides a tangible advantage for content moderation tasks within the fine-tuned SLM framework, enhancing both precision and recall metrics.

\section{Influence of Model Size}
\balance
\label{sec:appendix-sec-model-size}

Table \ref{tab:model-size-in-distribution-results} and Table \ref{tab:model-size-out-of-distribution-results} show detailed F1 scores for cross-task fine-tuned models of various sizes.

\begin{table*}\scriptsize
\centering
\begin{tabular}{llrrr}
\toprule
\multirow{2}{*}{Dataset}  & \multirow{2}{*}{Task}  & Phi-3-mini & Mixtral-8x7B & \multirow{2}{*}{\textbf{STAND-Guard}} \\
  &      & (CT-FT)    & (CT-FT)      &     \\
\midrule
PKU-Alignment BeaverTails   & Animal Abuse & 0.667      & 0.713  & 0.742      \\
 & Child Abuse  & 0.474      & 0.840  & 0.815      \\
  & Controversial Topics, Politics & 0.143      & 0.412  & 0.446      \\
  & Discrimination, Stereotype     & 0.674      & 0.747  & 0.731      \\
  & Drug Abuse, Weapons    & 0.687      & 0.744  & 0.746      \\
  & Financial \& Property Crime    & 0.720      & 0.765  & 0.744      \\
  & Hateful \& Offensive Language  & 0.676      & 0.671  & 0.670      \\
  & Misinformation   & 0.000      & 0.025  & 0.082      \\
  & Non-Violent Unethical Behavior & 0.589      & 0.683  & 0.655      \\
  & Privacy Violation      & 0.738      & 0.799  & 0.800      \\
  & Self Harm    & 0.593      & 0.710  & 0.727      \\
  & Sexually Explicit      & 0.547      & 0.653  & 0.667      \\
  & Terrorism, Organized Crime     & 0.045      & 0.125  & 0.196      \\
  & Violence     & 0.779      & 0.819  & 0.800      \\
PKU-Alignment   Safe-RLHF & Unsafe & 0.828      & 0.843  & 0.871      \\
Private & Hate & 0.734      & 0.796  & 0.827      \\
  & Self Harm    & 0.786      & 0.814  & 0.856      \\
  & Sexual & 0.696      & 0.808  & 0.802      \\
  & Violence     & 0.671      & 0.773  & 0.745      \\
\hline
\multicolumn{2}{c}{AVG}  & 0.581      & 0.671  & 0.680     
\\\bottomrule
\end{tabular} 
\caption{\label{tab:model-size-in-distribution-results}Impact of model size on F1 scores for fine-tuned models on in-distribution tasks under zero-shot setting. It is an expansion of Table~\ref{tab:f1-model-size}.}
\end{table*}

\begin{table*}
\scriptsize
\centering
\begin{tabular}{llrrr}
\toprule
\multirow{2}{*}{Dataset}  & \multirow{2}{*}{Task}   & Phi-3-mini    & Mixtral-8x7B & \multirow{2}{*}{\textbf{STAND-Guard}} \\
      & & (CT-FT) & (CT-FT)      &     \\
\midrule
Anatomy of Online Hate  & Hate  & 0.306   & 0.441  & 0.269      \\
BIG-bench (German)      & Gender Inclusive  & 0.000   & 0.677  & 0.708      \\
CallMeSexist    & Sexism  & 0.600   & 0.741  & 0.724      \\
Civil-Comments  & Insult  & 0.548   & 0.597  & 0.787      \\
      & Obscenity & 0.124   & 0.239  & 0.179      \\
      & Severe Toxicity & 0.458   & 0.363  & 0.530      \\
      & Sexually Explicit & 0.008   & 0.073  & 0.134      \\
      & Threat  & 0.088   & 0.127  & 0.163      \\
      & Toxicity  & 0.586   & 0.628  & 0.759      \\
Commonsense Morality    & Ethics  & 0.723   & 0.719  & 0.735      \\
COVID-HATE      & Hate  & 0.744   & 0.694  & 0.695      \\
CrowS-Pairs     & Bias  & 1.000   & 1.000  & 1.000      \\
DecodingTrust   & Stereotype      & 1.000   & 1.000  & 1.000      \\
DynaHate  & Hate  & 0.614   & 0.710  & 0.673      \\
Exaggerated Safety      & Safety  & 0.839   & 0.886  & 0.829      \\
GermEval (German) bin   & Offensive & 0.481   & 0.612  & 0.501      \\
GermEval (German) multi & Offensive & 0.511   & 0.677  & 0.101      \\
HASOC   (English) & Hate, Offensive & 0.655   & 0.643  & 0.679      \\
HASOC   (German)  & Hate, Offensive & 0.416   & 0.474  & 0.417      \\
Hate Speech and Offensive   Language      & Hate, Offensive & 0.837   & 0.719  & 0.585      \\
Hate Speech towards Foreigners   (German) & Hate  & 0.542   & 0.610  & 0.495      \\
HateCheck & Hate  & 0.856   & 0.925  & 0.867      \\
HateEval  & Hate  & 0.267   & 0.610  & 0.462      \\
HatemojiCheck   & Hate  & 0.705   & 0.879  & 0.836      \\
HateXplain      & Hate  & 0.761   & 0.801  & 0.782      \\
Jigsaw (Toxic Comment   Classification)   & Identity Hate   & 0.173   & 0.350  & 0.281      \\
      & Insult  & 0.395   & 0.494  & 0.416      \\
      & Obscene   & 0.422   & 0.574  & 0.512      \\
      & Severe Toxic    & 0.070   & 0.149  & 0.085      \\
      & Threat  & 0.063   & 0.328  & 0.300      \\
      & Toxic & 0.502   & 0.582  & 0.551      \\
Jiminy-Cricket  & Ethics  & 0.607   & 0.664  & 0.598      \\
Korean   Hate Speech (Korean)     & Hate  & 0.215   & 0.449  & 0.240      \\
      & Aggressiveness   & 0.430   & 0.510  & 0.382      \\
      & Bias  & 0.733   & 0.743  & 0.698      \\
OffComBR3 (Portuguese)  & Offensive & 0.426   & 0.555  & 0.558      \\
OpenAI CM & Harassment      & 0.384   & 0.494  & 0.726      \\
& Hateful   & 0.657   & 0.709  & 0.732      \\
& Self Harm & 0.731   & 0.712  & 0.928      \\
& Sexual  & 0.746   & 0.735  & 0.475      \\
& Violence  & 0.857   & 0.943  & 0.671      \\
PKU-Alignment-BeaverTails-Eval    & Unsafe  & 0.323   & 0.417  & 0.513      \\
Reddit Content Moderation   & Rule Moderation & 0.134   & 0.167  & 0.126      \\
RP-Mod \& RP-Crowd (German) & Offensive & 0.308   & 0.249  & 0.210      \\
Scruples Anecdotes      & Ethics  & 0.419   & 0.448  & 0.427      \\
Social   Bias Inference Corpus (SBIC)     & Intentionally Offensive & 0.717   & 0.721  & 0.709      \\
      & Potentially Offensive   & 0.715   & 0.740  & 0.728      \\
      & Sexually Offensive      & 0.464   & 0.515  & 0.195      \\
SWAD  & Swear & 0.597   & 0.627  & 0.539      \\
SWSR (Chinese) bin      & Sexism  & 0.433   & 0.559  & 0.504      \\
SWSR (Chinese) multi    & Sexism  & 0.520   & 0.538  & 0.528      \\
ToLD-BR (Portuguese)    & Offensive & 0.560   & 0.663  & 0.302      \\
      & Homophobia      & 0.187   & 0.375  & 0.140      \\
      & Misogyny  & 0.133   & 0.214  & 0.624      \\
      & Racism  & 0.152   & 0.325  & 0.215      \\
ToxiGen   & Toxic & 0.271   & 0.492  & 0.601      \\
TrustworthyLLM  & Safety  & 0.603   & 0.708  & 0.590      \\
TweetEval & Hate  & 0.615   & 0.602  & 0.556      \\
      & Irony & 0.766   & 0.758  & 0.685      \\
      & Offensive & 0.472   & 0.616  & 0.682      \\
USElectionHate  & Hate  & 0.222   & 0.487  & 0.392      \\
White Supremacist & Hate  & 0.540   & 0.733  & 0.739      \\
\hline
\multicolumn{2}{c}{AVG} & 0.488   & 0.577  & 0.533      
\\\bottomrule
\end{tabular}
\caption{\label{tab:model-size-out-of-distribution-results} Impact of model size on F1 scores for fine-tuned models on out-of-distribution tasks under zero-shot setting. It is an expansion of Table~\ref{tab:f1-model-size}.}
\end{table*}

%% file: coling.bbl
\begin{thebibliography}{64}
\providecommand{\natexlab}[1]{#1}

\bibitem[{Abdin et~al.(2024)Abdin, Jacobs, Awan, Aneja, Awadallah, Awadalla, Bach, Bahree, Bakhtiari, Behl et~al.}]{abdin2024phi}
Marah Abdin, Sam~Ade Jacobs, Ammar~Ahmad Awan, Jyoti Aneja, Ahmed Awadallah, Hany Awadalla, Nguyen Bach, Amit Bahree, Arash Bakhtiari, Harkirat Behl, et~al. 2024.
\newblock Phi-3 technical report: A highly capable language model locally on your phone.
\newblock \emph{arXiv preprint arXiv:2404.14219}.

\bibitem[{An et~al.(2023)An, Lin, Fu, Chen, Zheng, Lou, and Zhang}]{an2023context}
Shengnan An, Zeqi Lin, Qiang Fu, Bei Chen, Nanning Zheng, Jian-Guang Lou, and Dongmei Zhang. 2023.
\newblock How do in-context examples affect compositional generalization?
\newblock \emph{arXiv preprint arXiv:2305.04835}.

\bibitem[{Arora et~al.(2023)Arora, Nakov, Hardalov, Sarwar, Nayak, Dinkov, Zlatkova, Dent, Bhatawdekar, Bouchard et~al.}]{arora2023detecting}
Arnav Arora, Preslav Nakov, Momchil Hardalov, Sheikh~Muhammad Sarwar, Vibha Nayak, Yoan Dinkov, Dimitrina Zlatkova, Kyle Dent, Ameya Bhatawdekar, Guillaume Bouchard, et~al. 2023.
\newblock Detecting harmful content on online platforms: what platforms need vs. where research efforts go.
\newblock \emph{ACM Computing Surveys}, 56(3):1--17.

\bibitem[{Assenmacher et~al.(2021)Assenmacher, Niemann, M{\"u}ller, Seiler, Riehle, and Trautmann}]{Assenmacher2021RPModRPCrowdM}
Dennis Assenmacher, Marco Niemann, Kilian M{\"u}ller, Moritz~Vinzent Seiler, Dennis~M. Riehle, and Heike Trautmann. 2021.
\newblock \href {https://api.semanticscholar.org/CorpusID:244906720} {Rp-mod\&rp-crowd: Moderator- and crowd-annotated german news comment datasets}.
\newblock In \emph{NeurIPS Datasets and Benchmarks}.

\bibitem[{Barbieri et~al.(2020)Barbieri, Camacho-Collados, Neves, and Espinosa-Anke}]{barbieri2020tweeteval}
Francesco Barbieri, Jose Camacho-Collados, Leonardo Neves, and Luis Espinosa-Anke. 2020.
\newblock Tweeteval: Unified benchmark and comparative evaluation for tweet classification.
\newblock \emph{arXiv preprint arXiv:2010.12421}.

\bibitem[{Basile et~al.(2019)Basile, Bosco, Fersini, Nozza, Patti, Pardo, Rosso, and Sanguinetti}]{basile2019semeval}
Valerio Basile, Cristina Bosco, Elisabetta Fersini, Debora Nozza, Viviana Patti, Francisco Manuel~Rangel Pardo, Paolo Rosso, and Manuela Sanguinetti. 2019.
\newblock Semeval-2019 task 5: Multilingual detection of hate speech against immigrants and women in twitter.
\newblock In \emph{Proceedings of the 13th international workshop on semantic evaluation}, pages 54--63.

\bibitem[{bench authors(2023)}]{srivastava2023beyond}
BIG bench authors. 2023.
\newblock \href {https://openreview.net/forum?id=uyTL5Bvosj} {Beyond the imitation game: Quantifying and extrapolating the capabilities of language models}.
\newblock \emph{Transactions on Machine Learning Research}.

\bibitem[{Borkan et~al.(2019)Borkan, Dixon, Sorensen, Thain, and Vasserman}]{DBLP:journals/corr/abs-1903-04561}
Daniel Borkan, Lucas Dixon, Jeffrey Sorensen, Nithum Thain, and Lucy Vasserman. 2019.
\newblock \href {https://arxiv.org/abs/1903.04561} {Nuanced metrics for measuring unintended bias with real data for text classification}.
\newblock \emph{CoRR}, abs/1903.04561.

\bibitem[{Bretschneider and Peters(2017)}]{bretschneider2017detecting}
Uwe Bretschneider and Ralf Peters. 2017.
\newblock Detecting offensive statements towards foreigners in social media.

\bibitem[{Davidson et~al.(2017)Davidson, Warmsley, Macy, and Weber}]{davidson2017automated}
Thomas Davidson, Dana Warmsley, Michael Macy, and Ingmar Weber. 2017.
\newblock Automated hate speech detection and the problem of offensive language.
\newblock In \emph{Proceedings of the international AAAI conference on web and social media}, volume~11, pages 512--515.

\bibitem[{De~Gibert et~al.(2018)De~Gibert, Perez, Garc{\'\i}a-Pablos, and Cuadros}]{de2018hate}
Ona De~Gibert, Naiara Perez, Aitor Garc{\'\i}a-Pablos, and Montse Cuadros. 2018.
\newblock Hate speech dataset from a white supremacy forum.
\newblock \emph{arXiv preprint arXiv:1809.04444}.

\bibitem[{de~Gibert et~al.(2018)de~Gibert, Perez, Garc{\'\i}a-Pablos, and Cuadros}]{gibert2018hate}
Ona de~Gibert, Naiara Perez, Aitor Garc{\'\i}a-Pablos, and Montse Cuadros. 2018.
\newblock \href {https://doi.org/10.18653/v1/W18-5102} {{Hate Speech Dataset from a White Supremacy Forum}}.
\newblock In \emph{Proceedings of the 2nd Workshop on Abusive Language Online ({ALW}2)}, pages 11--20, Brussels, Belgium. Association for Computational Linguistics.

\bibitem[{de~Pelle and Moreira(2017)}]{Pelle2017}
Rogers~P. de~Pelle and Viviane~P. Moreira. 2017.
\newblock Offensive comments in the brazilian web: a dataset and baseline results.

\bibitem[{Dettmers et~al.(2024)Dettmers, Pagnoni, Holtzman, and Zettlemoyer}]{dettmers2024qlora}
Tim Dettmers, Artidoro Pagnoni, Ari Holtzman, and Luke Zettlemoyer. 2024.
\newblock Qlora: Efficient finetuning of quantized llms.
\newblock \emph{Advances in Neural Information Processing Systems}, 36.

\bibitem[{Franco et~al.(2023)Franco, Gaggi, and Palazzi}]{franco2023analyzing}
Mirko Franco, Ombretta Gaggi, and Claudio~E Palazzi. 2023.
\newblock Analyzing the use of large language models for content moderation with chatgpt examples.
\newblock In \emph{Proceedings of the 3rd International Workshop on Open Challenges in Online Social Networks}, pages 1--8.

\bibitem[{Ghosh et~al.(2024)Ghosh, Varshney, Galinkin, and Parisien}]{ghosh2024aegis}
Shaona Ghosh, Prasoon Varshney, Erick Galinkin, and Christopher Parisien. 2024.
\newblock Aegis: Online adaptive ai content safety moderation with ensemble of llm experts.
\newblock \emph{arXiv preprint arXiv:2404.05993}.

\bibitem[{Guo et~al.(2023)Guo, Hu, Mu, Shi, Zhao, Vishwamitra, and Hu}]{guo2023investigation}
Keyan Guo, Alexander Hu, Jaden Mu, Ziheng Shi, Ziming Zhao, Nishant Vishwamitra, and Hongxin Hu. 2023.
\newblock An investigation of large language models for real-world hate speech detection.
\newblock In \emph{2023 International Conference on Machine Learning and Applications (ICMLA)}, pages 1568--1573. IEEE.

\bibitem[{Hartvigsen et~al.(2022)Hartvigsen, Gabriel, Palangi, Sap, Ray, and Kamar}]{hartvigsen2022toxigen}
Thomas Hartvigsen, Saadia Gabriel, Hamid Palangi, Maarten Sap, Dipankar Ray, and Ece Kamar. 2022.
\newblock Toxigen: A large-scale machine-generated dataset for adversarial and implicit hate speech detection.
\newblock \emph{arXiv preprint arXiv:2203.09509}.

\bibitem[{He et~al.(2021)He, Ziems, Soni, Ramakrishnan, Yang, and Kumar}]{he2021racism}
Bing He, Caleb Ziems, Sandeep Soni, Naren Ramakrishnan, Diyi Yang, and Srijan Kumar. 2021.
\newblock Racism is a virus: Anti-asian hate and counterspeech in social media during the covid-19 crisis.
\newblock In \emph{Proceedings of the 2021 IEEE/ACM International Conference on Advances in Social Networks Analysis and Mining}, pages 90--94.

\bibitem[{He et~al.(2023)He, Zannettou, Shen, and Zhang}]{he2023you}
Xinlei He, Savvas Zannettou, Yun Shen, and Yang Zhang. 2023.
\newblock You only prompt once: On the capabilities of prompt learning on large language models to tackle toxic content.
\newblock In \emph{2024 IEEE Symposium on Security and Privacy (SP)}, pages 61--61. IEEE Computer Society.

\bibitem[{Hendrycks et~al.(2021{\natexlab{a}})Hendrycks, Burns, Basart, Critch, Li, Song, and Steinhardt}]{hendrycks2021aligning}
Dan Hendrycks, Collin Burns, Steven Basart, Andrew~Critch Critch, Jerry~Li Li, Dawn Song, and Jacob Steinhardt. 2021{\natexlab{a}}.
\newblock Aligning ai with shared human values.
\newblock In \emph{International Conference on Learning Representations}.

\bibitem[{Hendrycks et~al.(2021{\natexlab{b}})Hendrycks, Mazeika, Zou, Patel, Zhu, Navarro, Song, Li, and Steinhardt}]{hendrycks2021would}
Dan Hendrycks, Mantas Mazeika, Andy Zou, Sahil Patel, Christine Zhu, Jesus Navarro, Dawn Song, Bo~Li, and Jacob Steinhardt. 2021{\natexlab{b}}.
\newblock What would jiminy cricket do? towards agents that behave morally.
\newblock \emph{arXiv preprint arXiv:2110.13136}.

\bibitem[{Houlsby et~al.(2019)Houlsby, Giurgiu, Jastrzebski, Morrone, De~Laroussilhe, Gesmundo, Attariyan, and Gelly}]{houlsby2019parameter}
Neil Houlsby, Andrei Giurgiu, Stanislaw Jastrzebski, Bruna Morrone, Quentin De~Laroussilhe, Andrea Gesmundo, Mona Attariyan, and Sylvain Gelly. 2019.
\newblock Parameter-efficient transfer learning for nlp.
\newblock In \emph{International conference on machine learning}, pages 2790--2799. PMLR.

\bibitem[{Hu et~al.(2021)Hu, Wallis, Allen-Zhu, Li, Wang, Wang, Chen et~al.}]{hu2021lora}
Edward~J Hu, Phillip Wallis, Zeyuan Allen-Zhu, Yuanzhi Li, Shean Wang, Lu~Wang, Weizhu Chen, et~al. 2021.
\newblock Lora: Low-rank adaptation of large language models.
\newblock In \emph{International Conference on Learning Representations}.

\bibitem[{Hu et~al.(2022)Hu, Lee, Xie, Yu, Smith, and Ostendorf}]{hu2022context}
Yushi Hu, Chia-Hsuan Lee, Tianbao Xie, Tao Yu, Noah~A Smith, and Mari Ostendorf. 2022.
\newblock In-context learning for few-shot dialogue state tracking.
\newblock \emph{arXiv preprint arXiv:2203.08568}.

\bibitem[{Inan et~al.(2023)Inan, Upasani, Chi, Rungta, Iyer, Mao, Tontchev, Hu, Fuller, Testuggine et~al.}]{inan2023llama}
Hakan Inan, Kartikeya Upasani, Jianfeng Chi, Rashi Rungta, Krithika Iyer, Yuning Mao, Michael Tontchev, Qing Hu, Brian Fuller, Davide Testuggine, et~al. 2023.
\newblock Llama guard: Llm-based input-output safeguard for human-ai conversations.
\newblock \emph{arXiv preprint arXiv:2312.06674}.

\bibitem[{Ji et~al.(2024)Ji, Liu, Dai, Pan, Zhang, Bian, Chen, Sun, Wang, and Yang}]{ji2024beavertails}
Jiaming Ji, Mickel Liu, Josef Dai, Xuehai Pan, Chi Zhang, Ce~Bian, Boyuan Chen, Ruiyang Sun, Yizhou Wang, and Yaodong Yang. 2024.
\newblock Beavertails: Towards improved safety alignment of llm via a human-preference dataset.
\newblock \emph{Advances in Neural Information Processing Systems}, 36.

\bibitem[{Jiang et~al.(2022)Jiang, Yang, Liu, and Zubiaga}]{jiang2022swsr}
Aiqi Jiang, Xiaohan Yang, Yang Liu, and Arkaitz Zubiaga. 2022.
\newblock Swsr: A chinese dataset and lexicon for online sexism detection.
\newblock \emph{Online Social Networks and Media}, 27:100182.

\bibitem[{Jiang et~al.(2023)Jiang, Sablayrolles, Mensch, Bamford, Chaplot, Casas, Bressand, Lengyel, Lample, Saulnier et~al.}]{jiang2023mistral}
Albert~Q Jiang, Alexandre Sablayrolles, Arthur Mensch, Chris Bamford, Devendra~Singh Chaplot, Diego de~las Casas, Florian Bressand, Gianna Lengyel, Guillaume Lample, Lucile Saulnier, et~al. 2023.
\newblock Mistral 7b.
\newblock \emph{arXiv preprint arXiv:2310.06825}.

\bibitem[{Jiang et~al.(2024)Jiang, Sablayrolles, Roux, Mensch, Savary, Bamford, Chaplot, Casas, Hanna, Bressand et~al.}]{jiang2024mixtral}
Albert~Q Jiang, Alexandre Sablayrolles, Antoine Roux, Arthur Mensch, Blanche Savary, Chris Bamford, Devendra~Singh Chaplot, Diego de~las Casas, Emma~Bou Hanna, Florian Bressand, et~al. 2024.
\newblock Mixtral of experts.
\newblock \emph{arXiv preprint arXiv:2401.04088}.

\bibitem[{Kirk et~al.(2021)Kirk, Vidgen, R{\"o}ttger, Thrush, and Hale}]{kirk2021hatemoji}
Hannah~Rose Kirk, Bertram Vidgen, Paul R{\"o}ttger, Tristan Thrush, and Scott~A Hale. 2021.
\newblock Hatemoji: A test suite and adversarially-generated dataset for benchmarking and detecting emoji-based hate.
\newblock \emph{arXiv preprint arXiv:2108.05921}.

\bibitem[{Kolla et~al.(2024)Kolla, Salunkhe, Chandrasekharan, and Saha}]{kolla2024llm}
Mahi Kolla, Siddharth Salunkhe, Eshwar Chandrasekharan, and Koustuv Saha. 2024.
\newblock Llm-mod: Can large language models assist content moderation?
\newblock In \emph{Extended Abstracts of the CHI Conference on Human Factors in Computing Systems}, pages 1--8.

\bibitem[{Kumar et~al.(2024)Kumar, AbuHashem, and Durumeric}]{kumar2024watch}
Deepak Kumar, Yousef AbuHashem, and Zakir Durumeric. 2024.
\newblock Watch your language: Investigating content moderation with large language models.
\newblock \emph{arXiv preprint arXiv:2309.14517}.

\bibitem[{Leite et~al.(2020)Leite, Silva, Bontcheva, and Scarton}]{leite2020toxic}
Joao~A Leite, Diego~F Silva, Kalina Bontcheva, and Carolina Scarton. 2020.
\newblock Toxic language detection in social media for brazilian portuguese: New dataset and multilingual analysis.
\newblock \emph{arXiv preprint arXiv:2010.04543}.

\bibitem[{Lester et~al.(2021)Lester, Al-Rfou, and Constant}]{lester2021power}
Brian Lester, Rami Al-Rfou, and Noah Constant. 2021.
\newblock The power of scale for parameter-efficient prompt tuning.
\newblock In \emph{Proceedings of the 2021 Conference on Empirical Methods in Natural Language Processing}, pages 3045--3059.

\bibitem[{Li and Liang(2021)}]{li2021prefix}
Xiang~Lisa Li and Percy Liang. 2021.
\newblock Prefix-tuning: Optimizing continuous prompts for generation.
\newblock \emph{arXiv preprint arXiv:2101.00190}.

\bibitem[{Liu et~al.(2022)Liu, Ji, Fu, Tam, Du, Yang, and Tang}]{liu2022p}
Xiao Liu, Kaixuan Ji, Yicheng Fu, Weng Tam, Zhengxiao Du, Zhilin Yang, and Jie Tang. 2022.
\newblock P-tuning: Prompt tuning can be comparable to fine-tuning across scales and tasks.
\newblock In \emph{Proceedings of the 60th Annual Meeting of the Association for Computational Linguistics (Volume 2: Short Papers)}, pages 61--68.

\bibitem[{Liu et~al.(2023)Liu, Yao, Ton, Zhang, Cheng, Klochkov, Taufiq, and Li}]{liu2023trustworthy}
Yang Liu, Yuanshun Yao, Jean-Francois Ton, Xiaoying Zhang, Ruocheng Guo~Hao Cheng, Yegor Klochkov, Muhammad~Faaiz Taufiq, and Hang Li. 2023.
\newblock Trustworthy llms: a survey and guideline for evaluating large language models' alignment.
\newblock \emph{arXiv preprint arXiv:2308.05374}.

\bibitem[{Lourie et~al.(2020)Lourie, Bras, and Choi}]{Lourie2020Scruples}
Nicholas Lourie, Ronan~Le Bras, and Yejin Choi. 2020.
\newblock \href {https://arxiv.org/abs/2008.09094} {Scruples: A corpus of community ethical judgments on 32,000 real-life anecdotes}.
\newblock \emph{arXiv e-prints}.

\bibitem[{Ma et~al.(2023)Ma, Zhang, Fu, Zhao, and Wu}]{ma2023adapting}
Huan Ma, Changqing Zhang, Huazhu Fu, Peilin Zhao, and Bingzhe Wu. 2023.
\newblock Adapting large language models for content moderation: Pitfalls in data engineering and supervised fine-tuning.
\newblock \emph{arXiv preprint arXiv:2310.03400}.

\bibitem[{Mandl et~al.(2019)Mandl, Modha, Majumder, Patel, Dave, Mandlia, and Patel}]{mandl2019overview}
Thomas Mandl, Sandip Modha, Prasenjit Majumder, Daksh Patel, Mohana Dave, Chintak Mandlia, and Aditya Patel. 2019.
\newblock Overview of the hasoc track at fire 2019: Hate speech and offensive content identification in indo-european languages.
\newblock In \emph{Proceedings of the 11th annual meeting of the Forum for Information Retrieval Evaluation}, pages 14--17.

\bibitem[{Markov et~al.(2023)Markov, Zhang, Agarwal, Nekoul, Lee, Adler, Jiang, and Weng}]{markov2023holistic}
Todor Markov, Chong Zhang, Sandhini Agarwal, Florentine~Eloundou Nekoul, Theodore Lee, Steven Adler, Angela Jiang, and Lilian Weng. 2023.
\newblock A holistic approach to undesired content detection in the real world.
\newblock In \emph{Proceedings of the AAAI Conference on Artificial Intelligence}, volume~37, pages 15009--15018.

\bibitem[{Mathew et~al.(2021)Mathew, Saha, Yimam, Biemann, Goyal, and Mukherjee}]{mathew2021hatexplain}
Binny Mathew, Punyajoy Saha, Seid~Muhie Yimam, Chris Biemann, Pawan Goyal, and Animesh Mukherjee. 2021.
\newblock Hatexplain: A benchmark dataset for explainable hate speech detection.
\newblock In \emph{Proceedings of the AAAI conference on artificial intelligence}, volume~35, pages 14867--14875.

\bibitem[{Moon et~al.(2020)Moon, Cho, and Lee}]{moon2020beep}
Jihyung Moon, Won~Ik Cho, and Junbum Lee. 2020.
\newblock Beep! korean corpus of online news comments for toxic speech detection.
\newblock \emph{arXiv preprint arXiv:2005.12503}.

\bibitem[{Nangia et~al.(2020)Nangia, Vania, Bhalerao, and Bowman}]{nangia2020crows}
Nikita Nangia, Clara Vania, Rasika Bhalerao, and Samuel~R. Bowman. 2020.
\newblock {CrowS-Pairs: A Challenge Dataset for Measuring Social Biases in Masked Language Models}.
\newblock In \emph{Proceedings of the 2020 Conference on Empirical Methods in Natural Language Processing}, Online. Association for Computational Linguistics.

\bibitem[{Pamungkas et~al.(2020)Pamungkas, Basile, and Patti}]{pamungkas2020you}
Endang~Wahyu Pamungkas, Valerio Basile, and Viviana Patti. 2020.
\newblock Do you really want to hurt me? predicting abusive swearing in social media.
\newblock In \emph{Proceedings of the Twelfth Language Resources and Evaluation Conference}, pages 6237--6246.

\bibitem[{Qiao et~al.(2024)Qiao, Dogra, Stretcu, Lyu, Fang, Kwon, Lu, Luo, Wang, Chia et~al.}]{qiao2024scaling}
Wei Qiao, Tushar Dogra, Otilia Stretcu, Yu-Han Lyu, Tiantian Fang, Dongjin Kwon, Chun-Ta Lu, Enming Luo, Yuan Wang, Chih-Chun Chia, et~al. 2024.
\newblock Scaling up llm reviews for google ads content moderation.
\newblock In \emph{Proceedings of the 17th ACM International Conference on Web Search and Data Mining}, pages 1174--1175.

\bibitem[{R{\"o}ttger et~al.(2023)R{\"o}ttger, Kirk, Vidgen, Attanasio, Bianchi, and Hovy}]{rottger2023xstest}
Paul R{\"o}ttger, Hannah~Rose Kirk, Bertie Vidgen, Giuseppe Attanasio, Federico Bianchi, and Dirk Hovy. 2023.
\newblock Xstest: A test suite for identifying exaggerated safety behaviours in large language models.
\newblock \emph{arXiv preprint arXiv:2308.01263}.

\bibitem[{R{\"o}ttger et~al.(2020)R{\"o}ttger, Vidgen, Nguyen, Waseem, Margetts, and Pierrehumbert}]{rottger2020hatecheck}
Paul R{\"o}ttger, Bertram Vidgen, Dong Nguyen, Zeerak Waseem, Helen Margetts, and Janet~B Pierrehumbert. 2020.
\newblock Hatecheck: Functional tests for hate speech detection models.
\newblock \emph{arXiv preprint arXiv:2012.15606}.

\bibitem[{Salminen et~al.(2018)Salminen, Almerekhi, Milenkovi{\'c}, Jung, An, Kwak, and Jansen}]{salminen2018anatomy}
Joni Salminen, Hind Almerekhi, Milica Milenkovi{\'c}, Soon-gyo Jung, Jisun An, Haewoon Kwak, and Bernard Jansen. 2018.
\newblock Anatomy of online hate: developing a taxonomy and machine learning models for identifying and classifying hate in online news media.
\newblock In \emph{Proceedings of the International AAAI Conference on Web and Social Media}, volume~12.

\bibitem[{Samory et~al.(2021)Samory, Sen, Kohne, Fl{\"o}ck, and Wagner}]{samory2021call}
Mattia Samory, Indira Sen, Julian Kohne, Fabian Fl{\"o}ck, and Claudia Wagner. 2021.
\newblock “call me sexist, but...”: Revisiting sexism detection using psychological scales and adversarial samples.
\newblock In \emph{Proceedings of the international AAAI conference on web and social media}, volume~15, pages 573--584.

\bibitem[{Sap et~al.(2019)Sap, Gabriel, Qin, Jurafsky, Smith, and Choi}]{sap2019social}
Maarten Sap, Saadia Gabriel, Lianhui Qin, Dan Jurafsky, Noah~A Smith, and Yejin Choi. 2019.
\newblock Social bias frames: Reasoning about social and power implications of language.
\newblock \emph{arXiv preprint arXiv:1911.03891}.

\bibitem[{Sen et~al.(2024)Sen, Das, and Sen}]{sen2024hatetinyllm}
Tanmay Sen, Ansuman Das, and Mrinmay Sen. 2024.
\newblock Hatetinyllm: Hate speech detection using tiny large language models.
\newblock \emph{arXiv preprint arXiv:2405.01577}.

\bibitem[{Team et~al.(2024)Team, Mesnard, Hardin, Dadashi, Bhupatiraju, Pathak, Sifre, Rivi{\`e}re, Kale, Love et~al.}]{team2024gemma}
Gemma Team, Thomas Mesnard, Cassidy Hardin, Robert Dadashi, Surya Bhupatiraju, Shreya Pathak, Laurent Sifre, Morgane Rivi{\`e}re, Mihir~Sanjay Kale, Juliette Love, et~al. 2024.
\newblock Gemma: Open models based on gemini research and technology.
\newblock \emph{arXiv preprint arXiv:2403.08295}.

\bibitem[{Umansky et~al.()Umansky, Kubli, Donnay, Gilardi, Hangartner, Kotarcic, Bronner, Kurer, and Grech}]{umanskyenhancing}
Natalia Umansky, Ma{\"e}l Kubli, Karsten Donnay, Fabrizio Gilardi, Dominik Hangartner, Ana Kotarcic, Laura Bronner, Selina Kurer, and Philip Grech.
\newblock Enhancing hate speech detection with fine-tuned large language models requires high-quality data.

\bibitem[{USElectionHate()}]{USElectionHate}
USElectionHate.
\newblock \emph{https://www.ims.uni-stuttgart.de/forschung/ressourcen/korpora/stance-hof/}.

\bibitem[{Vidgen et~al.(2021)Vidgen, Thrush, Waseem, and Kiela}]{vidgen2021learning}
Bertie Vidgen, Tristan Thrush, Zeerak Waseem, and Douwe Kiela. 2021.
\newblock Learning from the worst: Dynamically generated datasets to improve online hate detection.
\newblock In \emph{Proceedings of the 59th Annual Meeting of the Association for Computational Linguistics and the 11th International Joint Conference on Natural Language Processing (Volume 1: Long Papers)}, pages 1667--1682.

\bibitem[{Wang et~al.(2023{\natexlab{a}})Wang, Chen, Pei, Xie, Kang, Zhang, Xu, Xiong, Dutta, Schaeffer et~al.}]{wang2023decodingtrust}
Boxin Wang, Weixin Chen, Hengzhi Pei, Chulin Xie, Mintong Kang, Chenhui Zhang, Chejian Xu, Zidi Xiong, Ritik Dutta, Rylan Schaeffer, et~al. 2023{\natexlab{a}}.
\newblock Decodingtrust: A comprehensive assessment of trustworthiness in gpt models.
\newblock \emph{arXiv preprint arXiv:2306.11698}.

\bibitem[{Wang et~al.(2023{\natexlab{b}})Wang, Li, Han, Nakov, and Baldwin}]{wang2023not}
Yuxia Wang, Haonan Li, Xudong Han, Preslav Nakov, and Timothy Baldwin. 2023{\natexlab{b}}.
\newblock Do-not-answer: A dataset for evaluating safeguards in llms.
\newblock \emph{arXiv preprint arXiv:2308.13387}.

\bibitem[{Wiegand et~al.(2018)Wiegand, Siegel, and Ruppenhofer}]{wiegand2018overview}
Michael Wiegand, Melanie Siegel, and Josef Ruppenhofer. 2018.
\newblock Overview of the germeval 2018 shared task on the identification of offensive language.

\bibitem[{Wullach et~al.(2021)Wullach, Adler, and Minkov}]{wullach2021fight}
Tomer Wullach, Amir Adler, and Einat Minkov. 2021.
\newblock Fight fire with fire: Fine-tuning hate detectors using large samples of generated hate speech.
\newblock In \emph{Findings of the Association for Computational Linguistics: EMNLP 2021}, pages 4699--4705.

\bibitem[{Yuan et~al.(2024)Yuan, Xiong, Zeng, Yu, Jia, Song, and Li}]{yuan2024rigorllm}
Zhuowen Yuan, Zidi Xiong, Yi~Zeng, Ning Yu, Ruoxi Jia, Dawn Song, and Bo~Li. 2024.
\newblock Rigorllm: Resilient guardrails for large language models against undesired content.
\newblock \emph{arXiv preprint arXiv:2403.13031}.

\bibitem[{Zhang et~al.(2024)Zhang, Wu, Xu, Cao, Du, and Psounis}]{zhang2024efficient}
Jiang Zhang, Qiong Wu, Yiming Xu, Cheng Cao, Zheng Du, and Konstantinos Psounis. 2024.
\newblock Efficient toxic content detection by bootstrapping and distilling large language models.
\newblock In \emph{Proceedings of the AAAI Conference on Artificial Intelligence}, volume~38, pages 21779--21787.

\bibitem[{Zhang et~al.(2023)Zhang, Luo, Chuang, Fang, Gaitskell, Hartvigsen, Wu, Fox, Meng, and Glass}]{zhang2023interpretable}
Tianhua Zhang, Hongyin Luo, Yung-Sung Chuang, Wei Fang, Luc Gaitskell, Thomas Hartvigsen, Xixin Wu, Danny Fox, Helen Meng, and James Glass. 2023.
\newblock Interpretable unified language checking.
\newblock \emph{arXiv preprint arXiv:2304.03728}.

\end{thebibliography}
